\documentclass[10pt,journal,compsoc]{IEEEtran}
\usepackage[utf8]{inputenc}
\usepackage{booktabs}
\usepackage{graphicx}
\usepackage[caption=false,font=normalsize,labelfont=sf,textfont=sf]{subfig}
\usepackage{amsmath,amssymb,amsthm,mathtools}
\usepackage{multirow,array,tabularx}
\usepackage[nocompress]{cite}
\usepackage[hidelinks]{hyperref}
\newcolumntype{L}{>{\raggedright\arraybackslash}X}
\newcommand{\thead}[1]{\shortstack[c]{#1}}
\newtheorem{theorem}{Theorem}
\newtheorem{corollary}{Corollary}

\begin{document}

\title{Few-Shot Cross-Subject Neural--Image Retrieval via Stimulus-Anchored Latent Bias Calibration}

\author{Xizhang Yao, Li Zhang, Gan Huang, and Zhen Liang%
\IEEEcompsocitemizethanks{\IEEEcompsocthanksitem
The authors are with the School of Biomedical Engineering, Health Science Center,
Shenzhen University, and the Guangdong Provincial Key Laboratory of Biomedical
Measurements and Ultrasound Imaging, Shenzhen, China.
E-mail: yaoxizhang2022@email.szu.edu.cn; lzhang@szu.edu.cn;
huanggan@szu.edu.cn; janezliang@szu.edu.cn.}}

\IEEEtitleabstractindextext{
\begin{abstract}
In cross-subject neural--image retrieval, comparing the mean of a few stimulus-matched target observations with the global source mean conflates subject-specific representation shifts with stimulus-composition effects. To address this problem, we propose Stimulus-Anchored Latent Bias Calibration (SA-LBC), which estimates the first-order shift under fixed preprocessing and a frozen source-trained representation by averaging residuals between target responses and same-stimulus source-population anchors, then shrinks the correction using source-derived uncertainty. Under uniform sampling without replacement from a finite stimulus population, we derive exact finite-population MSE expressions for the unshrunk anchored and global-mean estimators, including residual--anchor cross-covariance, and establish a necessary and sufficient condition for anchoring to reduce MSE relative to global-mean estimation. In a 10-subject leave-one-subject-out evaluation on THINGS-EEG2, SA-LBC increases mean reciprocal rank (MRR) for the Attention--CLIP ViT-B/32 configuration from 0.1867 without calibration to 0.2013 and 0.2336 with $M=2$ and $M=16$ stimulus-level calibration samples, respectively. Across six neural--visual configurations and all calibration budgets, stimulus-matched anchors yield higher MRR than anchors with shuffled stimulus identities, but retrieval gains over no calibration remain representation-dependent. At (M=2), shrinkage improves MRR over the unshrunk anchored estimator in all six configurations, but its effects at larger budgets vary across representations; the relative ranking of translation and orientation corrections also depends on regularization strength. Thus, lower shift-estimation MSE does not guarantee higher retrieval performance.
The source code is publicly available at:
\url{https://github.com/Xizhang-Yao/SA-LBC}.
\end{abstract}

}

\maketitle
\IEEEdisplaynontitleabstractindextext

\IEEEpeerreviewmaketitle

\section{Introduction}

Non-invasive neural--image retrieval aims to identify visual content from brain activity by aligning neural and visual representations in a shared space~\cite{songNICE2024,chenMindsEyeImage2024,wangNeuroCLIPBrainInspiredPrompt2025a}.
Recent EEG- and MEG-based methods have substantially advanced neural--image retrieval under subject-dependent protocols, with models trained and evaluated separately for each participant~\cite{wangNeuroCLIPBrainInspiredPrompt2025a,jiangSubjectAwareMultiGranularityAlignment2026,joHyFIHyperbolicFeature2026}.
However, the learned neural--visual correspondence often transfers poorly to unseen subjects, partly because neural responses exhibit systematic inter-subject variability~\cite{liCrossSubjectGeneralizationEEG2026a,cuiSCORESubjectCoordinate2026}.
Reliable transfer of such correspondence to new subjects therefore remains a key obstacle to scalable cross-subject deployment of neural--image retrieval.

Zero-shot cross-subject generalization avoids target-specific calibration
but cannot use target observations to correct individual
mismatch~\cite{choiSubjectInvariantEEGEmbeddings2026,huangNEEDCrossSubjectCrossTask2025,kimCroSSCrosssubjectVisual2026b}.
In contrast, adaptation with substantial new-subject data can accommodate
individual variability but incurs additional acquisition and adaptation
costs~\cite{ferranteTheirEyesMultisubject2024a,scottiMindEye2SharedSubjectModels2024}.
A complementary setting uses a few stimulus-matched target observations to estimate a latent correction before testing, conditional on fixed preprocessing and a frozen source-trained representation.
Here, shared stimulus identities link target observations to source
responses elicited by the same stimuli, providing a reference for
estimating subject
mismatch~\cite{haxbyCommonHighDimensionalModel2011,ferranteTheirEyesMultisubject2024a,liuMindAdapterFewShotParameterEfficient2026}.
These correspondences specify what to compare, but not which
cross-subject corrections can be estimated reliably from so few
observations.

Candidate corrections range from translation and scaling to orientation changes and nonlinear mappings~\cite{rodriguesRiemannianProcrustesAnalysis2019a,liuMindAdapterFewShotParameterEfficient2026}.
Positive global isotropic scaling leaves cosine-based retrieval rankings unchanged.
Translation and orientation offer two contrasting cases: translation uses centroid differences, whereas orientation uses the geometry of matched, centered responses~\cite{rodriguesRiemannianProcrustesAnalysis2019a,cuiSCORESubjectCoordinate2026}.
With \(M\) matched observations in a \(d\)-dimensional space, the centered cross-covariance has rank at most \(M-1\), limiting its data-supported modes when \(M\ll d\).
Translation avoids estimating this cross-covariance, but a small target-sample mean and a global source mean need not represent the same stimuli.
Their difference therefore reflects sampled-stimulus composition as well as subject mismatch, creating \emph{stimulus-composition uncertainty} even for first-order calibration.

\begin{figure*}[!t]
    \centering
    \includegraphics[width=\textwidth]{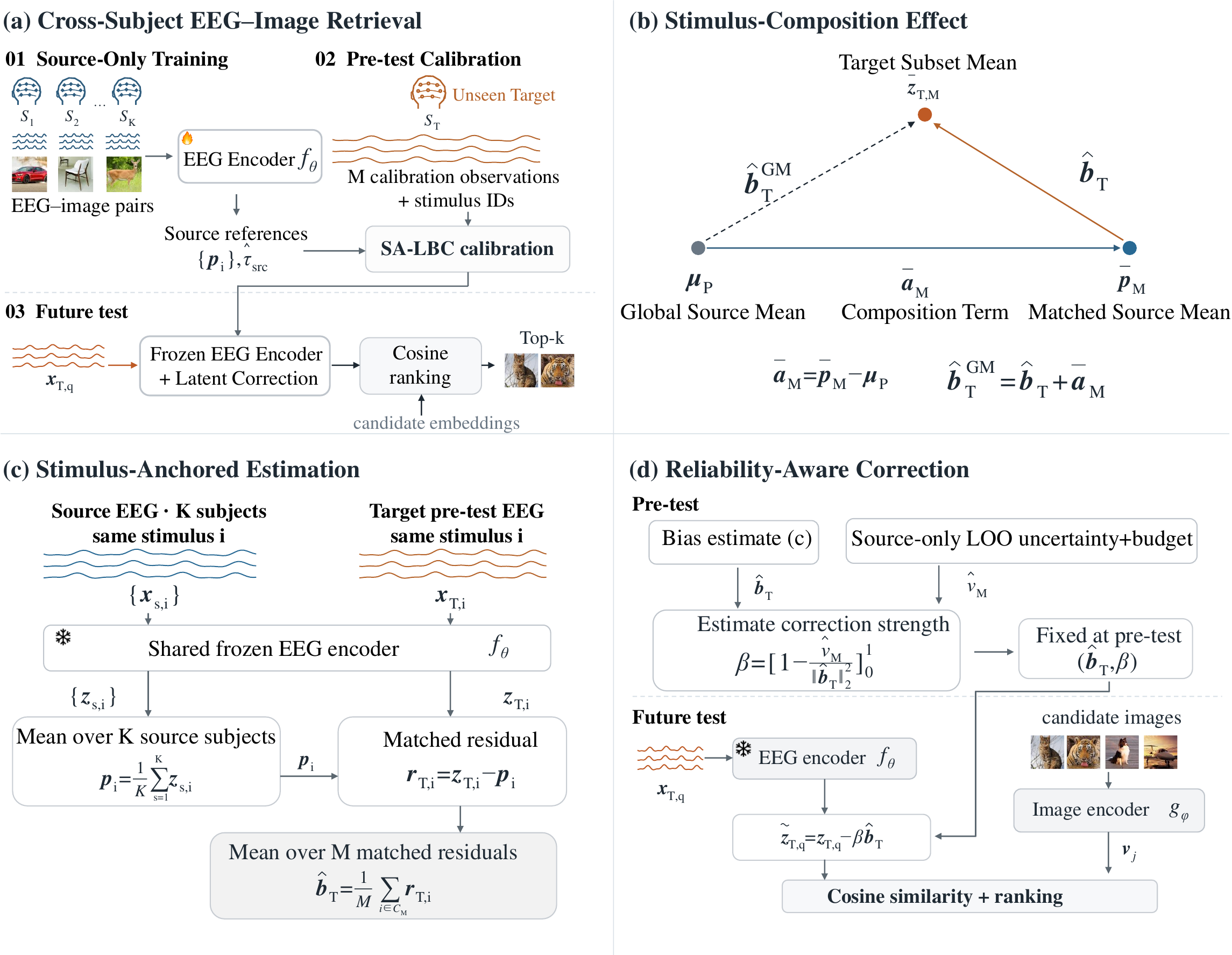}
    \caption{
    Overview of Stimulus-Anchored Latent Bias Calibration (SA-LBC),
    illustrated for EEG--image retrieval after fixed preprocessing.
    (a) Source-only representation training is followed by calibration from \(M\) pre-test target observations and retrieval on concept-disjoint TEST stimuli.
    (b) The global-mean and anchored estimates differ by the sampled-stimulus composition term \(\bar{\mathbf a}_M = \bar{\mathbf p}_M - \boldsymbol{\mu}_P\).
    (c) Source embeddings are averaged across subjects for each stimulus to form population anchors; averaging target--anchor residuals yields the shift estimate \(\widehat{\mathbf b}_T\).
    (d) Source-only leave-one-out (LOO) residual uncertainty, the sampling budget, and the estimated shift energy determine the plug-in shrinkage factor \(\beta\).
    The fixed correction is applied before \(\ell_2\) normalization and cosine ranking.
    Stimulus IDs select the source anchors; latent calibration uses no image pixels or visual features, target TEST statistics, or retrieval feedback, and does not update the representation.
    In THINGS-EEG2, each calibration observation averages four responses, and participant-specific TRAIN whitening uses additional target recordings outside the \(M\)-observation budget.
    }
    \label{fig:framework}
\end{figure*}

To address this uncertainty, we introduce Stimulus-Anchored Latent Bias Calibration (SA-LBC), which estimates a target-specific first-order shift using stimulus-matched source-population anchors (Fig.~\ref{fig:framework}).
SA-LBC compares the target calibration mean with the source-population mean for the same stimulus set, equivalently averaging the matched residuals.
This removes the explicit sampled-stimulus composition term incurred in global-mean matching, but finite-sample residual variation remains.
SA-LBC therefore applies reliability-aware shrinkage, attenuating poorly supported shifts rather than always applying them in full.
To isolate anchoring from shrinkage, we analyze the unshrunk anchored and global-mean estimators under uniform sampling without replacement from a finite stimulus population.
This yields exact mean-squared errors and a necessary and sufficient condition for anchoring to reduce estimation error.
Finally, same-budget comparisons of translation, centroid-preserving regularized orientation, and their combination test whether paired geometry adds retrieval value.

Under leave-one-subject-out evaluation with source-only model selection and the stated preprocessing, SA-LBC improves MRR from 0.1867 to 0.2013 with \(M=2\) stimulus-level calibration averages and to 0.2336 with \(M=16\) in the Attention--CLIP ViT-B/32 configuration on THINGS-EEG2.
Across six neural--visual configurations, mean relative discrepancies between analytic and empirical MSE are below 1.7\% for both unshrunk estimators.
Matched source references outperform references constructed with shuffled stimulus identities in all six configurations, but calibration does not consistently improve on leaving the representation unchanged.
The benefit of shrinkage varies across representations, and geometric ordering can reverse with regularization strength.
These results distinguish estimating the shift, controlling its strength, and establishing its retrieval utility.

Our contributions are threefold:
\begin{itemize}
    \item We introduce SA-LBC, a few-shot pre-test latent-calibration method that combines stimulus-matched source-population references with source-derived shrinkage in a frozen neural--visual representation.
    \item We derive exact finite-population MSE expressions for the unshrunk anchored and global-mean estimators, retaining residual--anchor covariance to establish when anchoring lowers estimation error.
    \item We characterize the benefits and limits of source-reference matching, shrinkage, and regularized geometric correction through same-budget comparisons across six neural--visual configurations.
\end{itemize}

\section{Related Work}
Prior work relevant to cross-subject neural--image retrieval can be understood through two questions: how neural--visual representations are learned, and what target-subject information is available when transferring them to a new individual.
We therefore review neural--image representation learning, cross-subject alignment and source-side generalization, unlabeled target-side calibration, and limited-data cross-subject adaptation.
\subsection{Neural--Image Retrieval}
Early visual brain-decoding studies established that perceptual content can be identified or reconstructed from neural activity~\cite{kamitani2005decoding,kay2008identifying,naselaris2009bayesian,horikawaGenericDecodingSeen2017}.
Later work adopted multimodal representation learning, embedding neural activity and visual content into directly comparable feature spaces~\cite{palazzoDecodingBrainRepresentations2021,liVisualDecodingReconstruction2024a,chenMindsEyeImage2024,liNeuralMCRLNeuralMultimodal2024}.
Recent EEG-to-image retrieval methods extend this formulation by aligning EEG with pretrained visual representations in a shared embedding space for similarity-based retrieval~\cite{wangNeuroCLIPBrainInspiredPrompt2025a,zhangNeuroBridgeBioInspiredSelfSupervised2025,zhouEEGiTTeachingVision,xuNeurovistaBidirectionalMasked2026}.

One recent direction refines where and at what level neural and visual information should be matched, moving from a single global embedding toward multi-granularity, hierarchical, regional, and structured correspondences~\cite{jiangSubjectAwareMultiGranularityAlignment2026,joHyFIHyperbolicFeature2026,tangAligningWhatEEG2026,zhuRegionAwareHierarchicalSubFeature}.
A second direction enriches what visual information is used for alignment, incorporating semantic priors, complementary visual teachers, and saliency- or object-aware supervision~\cite{liuLinguisticPriorsVisual2026,wangCORTIVACandidateScoreFusion2026,linSIMONSaliencyawareIntegrative2026}.
These advances improve how neural and visual information is represented and aligned, but do not by themselves determine how the resulting representation should be transferred to an unseen subject.

\subsection{Cross-Subject Alignment and Source-Side Generalization}

Cross-subject variability has commonly been addressed in two ways: by explicitly aligning subject-specific representations or by learning representations that generalize across subjects during source training.
The first strategy estimates mappings between subject-specific representational spaces using cross-subject correspondence or shared structure~\cite{haxbyCommonHighDimensionalModel2011,rodriguesRiemannianProcrustesAnalysis2019a}.
Geometric formulations make such mappings explicit through changes in location, scale, and orientation~\cite{rodriguesRiemannianProcrustesAnalysis2019a}.
Related alignment methods also standardize EEG trials in Euclidean
space~\cite{he2020euclidean} or match source and target feature
covariances~\cite{sun2016coral}.
The second strategy shifts the burden to source training, seeking representations that transfer across individuals through feature alignment, invariance learning, disentanglement, and related domain-generalization objectives~\cite{liCrossSubjectGeneralizationEEG2026a,luCausalityDrivenConvolutionalManifold2026a}.
This strategy has been pursued through subject-aware normalization, adversarially learned subject-invariant embeddings, and cross-subject reconstruction frameworks designed for unseen individuals~\cite{kimCroSSCrosssubjectVisual2026b,choiSubjectInvariantEEGEmbeddings2026,huangNEEDCrossSubjectCrossTask2025}.

Together, these approaches reduce subject variability either by explicitly aligning subject-specific representations or by learning representations that generalize across subjects.
Yet dedicated benchmarks and empirical studies still show a substantial gap between cross-subject and within-subject decoding~\cite{zhuCrossPTEEGBenchmarkCrossParticipant2024,wangWhatCausesPerformance2026,liCrossSubjectGeneralizationEEG2026a}.
This motivates a complementary deployment-time question: given a trained representation, how can target-subject observations be used to correct the mismatch that remains?

\subsection{Unlabeled Target-Side Calibration}

Once a neural--visual representation has been trained, unlabeled target-side information can still be used to correct unseen-subject mismatch at deployment~\cite{huangSATTCStructureAwareLabelFree,cuiSCORESubjectCoordinate2026}.
SATTC follows this strategy by keeping the EEG and image encoders fixed and recalibrating retrieval from the unlabeled target test batch through statistical normalization and structure-aware similarity correction~\cite{huangSATTCStructureAwareLabelFree}.
SCORE takes a more explicit geometric view, recovering target coordinates at deployment from unlabeled target observations through moment matching, pseudo-landmark discovery, and regularized orthogonal alignment~\cite{cuiSCORESubjectCoordinate2026}.

Their target information, however, takes different forms: SATTC is naturally transductive over the unlabeled test batch, whereas SCORE can estimate a coordinate map from unlabeled target observations and apply the frozen map to a disjoint unseen gallery~\cite{cuiSCORESubjectCoordinate2026}.
Both methods infer target mismatch from unlabeled target-side structure, without requiring explicit stimulus correspondence across subjects.
A different regime arises when a small set of stimulus-matched observations is available before evaluation: each target observation can then be directly related to the source representation of the same stimulus.
This correspondence changes the calibration problem from inferring mismatch from target-side structure to estimating it from paired cross-subject evidence.
Such settings motivate limited-data cross-subject adaptation methods, which we review next.

\subsection{Limited-Data Cross-Subject Adaptation}

Limited-data cross-subject adaptation differs in where the target-specific correction is introduced.
One line estimates an explicit transformation from the new subject to a source or common representational space, using shared or related stimuli to establish cross-subject correspondence~\cite{ferranteTheirEyesMultisubject2024a,daiMindAlignerExplicitBrain2025}.
Another adapts the pretrained decoding model or a lightweight subject-specific module using limited target data, as in MindShot~\cite{jiangMindShotFewshotBrain2026}, ENIGMA~\cite{kneelandENIGMAEEGtoImage152026}, and MindAdapter~\cite{liuMindAdapterFewShotParameterEfficient2026}.

Together, these studies show that limited target data can support cross-subject transfer through functional mapping, model adaptation, or residual correction.
They do not, however, directly ask which subject-specific correction can be estimated reliably under an extremely small, fixed budget of stimulus-matched pre-test observations.
Our work treats this question as an explicit estimation problem.

\section{Method}\label{sec:method}

\subsection{Few-Shot Pre-Test Calibration Problem}

We consider \(K\) source subjects
\(\mathcal S=\{1,\ldots,K\}\)
and an unseen target subject \(T\) that is excluded from source-model training.
Let \(f_{\theta}\) be a pretrained frozen neural encoder, and let
\(\mathbf x_{s,i}\) and \(\mathbf x_{T,i}\) denote the preprocessed neural observations of source subject \(s\) and target subject \(T\) for stimulus \(i\), respectively.
Their latent representations are
\begin{equation}
    \mathbf z_{s,i}
    =
    f_{\theta}(\mathbf x_{s,i}),
    \qquad
    \mathbf z_{T,i}
    =
    f_{\theta}(\mathbf x_{T,i}),
    \qquad
    \mathbf z_{s,i},\mathbf z_{T,i}
    \in\mathbb R^{d}.
    \label{eq:latent_embeddings}
\end{equation}
Let \(g_{\phi}\) be the frozen visual encoder, with candidate-image representation
\begin{equation}
    \mathbf v_j
    =
    g_{\phi}(\mathbf I_j)
    \in\mathbb R^d.
\end{equation}
After calibration, candidate images are ranked by cosine similarity between the calibrated target representation and the frozen visual representations.

The budget below applies to latent calibration after preprocessing.
In THINGS-EEG2, whitening is participant specific and uses each participant's TRAIN recordings, including those of the held-out target.
These preprocessing observations are outside the \(M\)-observation calibration budget.
LOSO excludes the target from representation fitting and checkpoint selection, not from participant-specific preprocessing.

Let the pre-test calibration pool for target subject \(T\) contain \(N_T\) stimuli.
Only a subset
\begin{equation}
    \mathcal C_M
    \subseteq
    \{1,\ldots,N_T\},
    \qquad
    |\mathcal C_M|=M,
    \qquad
    M\ll N_T
    \label{eq:calibration_set}
\end{equation}
is available for estimating the latent correction.
For each calibration observation, the stimulus identity is used only to index the corresponding source-population neural anchor; the calibration image, its visual embedding, and semantic category labels are not used.
The calibration pool is concept-disjoint from the final TEST set, and calibration uses neither target TEST statistics nor retrieval feedback and performs no target-side parameter updates.
The calibration task is to estimate a subject-specific latent correction from \(\mathcal C_M\) and apply it to future target observations while keeping the neural--visual representation fixed (Fig.~\ref{fig:framework}(a)).

\subsection{Stimulus-Composition Uncertainty in First-Order Shift Estimation}

We first consider a first-order correction that aligns the target calibration centroid with the source-population centroid.
The target calibration centroid is
\begin{equation}
    \bar{\mathbf z}_{T,M}
    =
    \frac{1}{M}
    \sum_{i\in\mathcal C_M}
    \mathbf z_{T,i}.
    \label{eq:target_calibration_mean}
\end{equation}
The source-population centroid over the same \(N_T\)-stimulus population is
\begin{equation}
    \boldsymbol{\mu}_P
    =
    \frac{1}{K N_T}
    \sum_{s=1}^{K}
    \sum_{i=1}^{N_T}
    \mathbf z_{s,i}.
    \label{eq:population_global_mean}
\end{equation}
Global-mean matching therefore estimates the target-specific shift as
\begin{equation}
    \widehat{\mathbf b}_{T}^{\mathrm{GM}}
    =
    \bar{\mathbf z}_{T,M}
    -
    \boldsymbol{\mu}_P.
    \label{eq:global_mean_estimator}
\end{equation}

Because \(\bar{\mathbf z}_{T,M}\) averages only the sampled calibration stimuli, it depends on both the target subject and the composition of \(\mathcal C_M\).
Consequently, two calibration subsets of the same size can yield different global-mean estimates for the same target subject because they contain different stimuli.
We refer to this finite-sample contribution to first-order estimation error as \emph{stimulus-composition uncertainty}.
This formulation isolates a finite-sample estimation problem within first-order calibration; it does not assume that all cross-subject mismatch is additive.

\subsection{Stimulus-Matched Source-Population Anchors}

To match the sampled stimulus composition on the source side, we construct a stimulus-specific source-population anchor by averaging the frozen representations of stimulus \(i\) across source subjects,
\begin{equation}
    \mathbf p_i
    =
    \frac{1}{K}
    \sum_{s=1}^{K}
    \mathbf z_{s,i}.
    \label{eq:population_anchor}
\end{equation}
The resulting anchor is the source-population mean latent representation of stimulus \(i\).
The global source-population centroid is therefore
\begin{equation}
    \boldsymbol{\mu}_P
    =
    \frac{1}{N_T}
    \sum_{i=1}^{N_T}
    \mathbf p_i.
    \label{eq:anchor_global_mean}
\end{equation}
For the sampled calibration stimuli, the matched source-population centroid is
\begin{equation}
    \bar{\mathbf p}_M
    =
    \frac{1}{M}
    \sum_{i\in\mathcal C_M}
    \mathbf p_i.
    \label{eq:matched_anchor_mean}
\end{equation}
Because \(\bar{\mathbf z}_{T,M}\) and \(\bar{\mathbf p}_M\) average exactly the same stimulus identities, their difference cancels the explicit source-population composition term introduced by global-mean matching (Fig.~\ref{fig:framework}(b)).

\subsection{Stimulus-Anchored Latent Bias Calibration}

With a source-population anchor matched to each calibration stimulus, we define the stimulus-wise residual
\begin{equation}
    \mathbf r_{T,i}
    =
    \mathbf z_{T,i}
    -
    \mathbf p_i.
    \label{eq:stimulus_matched_residual}
\end{equation}
SA-LBC estimates the target-specific first-order shift by averaging these residuals (Fig.~\ref{fig:framework}(c)),
\begin{align}
    \widehat{\mathbf b}_{T}^{\mathrm{SA}}
    &=
    \frac{1}{M}
    \sum_{i\in\mathcal C_M}
    \left(
    \mathbf z_{T,i}
    -
    \mathbf p_i
    \right)
    \nonumber\\
    &=
    \bar{\mathbf z}_{T,M}
    -
    \bar{\mathbf p}_M.
    \label{eq:salbc_estimator}
\end{align}
The translation estimate depends on the target and anchor means,
not on the pairing order within the selected sets.
Any permutation of the selected anchors leaves the estimate unchanged.
Stimulus identities are therefore used to construct a source reference
with the same stimulus composition as the target calibration set.
Replacing the global source centroid
\(\boldsymbol{\mu}_P\)
with the subset-matched centroid
\(\bar{\mathbf p}_M\)
removes the explicit source-population composition term identified above.
The estimator nevertheless remains variable across finite calibration subsets because the matched residuals themselves vary across stimuli.
Henceforth, we write
\(\widehat{\mathbf b}_{T}\equiv\widehat{\mathbf b}_{T}^{\mathrm{SA}}\).

\subsection{Reliability-Aware Shrinkage}

The correction below is a source-derived plug-in shrinkage rule,
related in form to classical shrinkage estimation~\cite{james1961quadratic}.
To quantify the remaining finite-sample uncertainty without target TEST information, we treat each source subject in turn as a pseudo-target.
For source subject \(s\), we form a subject-wise leave-one-out (LOO) anchor by excluding \(s\) from the source population,
\begin{equation}
    \mathbf p_i^{(-s)}
    =
    \frac{1}{K-1}
    \sum_{\substack{q\in\mathcal S\\q\neq s}}
    \mathbf z_{q,i}.
    \label{eq:loo_anchor}
\end{equation}
The corresponding LOO residual is
\begin{equation}
    \mathbf e_{s,i}^{\mathrm{LOO}}
    =
    \mathbf z_{s,i}
    -
    \mathbf p_i^{(-s)}.
    \label{eq:loo_residual}
\end{equation}
Its mean over the \(N_T\)-stimulus population is
\begin{equation}
    \bar{\mathbf e}_{s}^{\mathrm{LOO}}
    =
    \frac{1}{N_T}
    \sum_{i=1}^{N_T}
    \mathbf e_{s,i}^{\mathrm{LOO}},
    \label{eq:loo_residual_mean}
\end{equation}
and the across-stimulus residual covariance trace is estimated by
\begin{equation}
    \widehat{\tau}_{s}^{\mathrm{LOO}}
    =
    \frac{1}{N_T-1}
    \sum_{i=1}^{N_T}
    \left\|
    \mathbf e_{s,i}^{\mathrm{LOO}}
    -
    \bar{\mathbf e}_{s}^{\mathrm{LOO}}
    \right\|_2^2.
    \label{eq:loo_noise_trace}
\end{equation}

The LOO residual uses an anchor averaged over \(K-1\) subjects, whereas the deployed target residual uses an anchor averaged over all \(K\) source subjects.
Under a working approximation of comparable subject-level residual covariance and negligible cross-subject covariance, subtracting an \(n\)-subject mean scales the covariance trace by approximately \(1+1/n\).
We therefore estimate the target-residual covariance trace as
\begin{equation}
    \widehat{\tau}_{\mathrm{src}}
    =
    \frac{1+1/K}
         {1+1/(K-1)}
    \cdot
    \frac{1}{K}
    \sum_{s=1}^{K}
    \widehat{\tau}_{s}^{\mathrm{LOO}}.
    \label{eq:source_noise_energy}
\end{equation}
This variance-transfer approximation is used only for deployment-time shrinkage; the exact finite-population analysis below does not rely on it.

For a size-\(M\) calibration subset sampled without replacement from the \(N_T\)-stimulus population, the estimated sampling-error energy of the residual mean is
\begin{equation}
    \widehat{\nu}_M
    =
    \widehat{\tau}_{\mathrm{src}}
    \left(
    \frac{1}{M}
    -
    \frac{1}{N_T}
    \right).
    \label{eq:finite_population_noise}
\end{equation}
Under a signal-plus-sampling-error view,
\(\|\widehat{\mathbf b}_T\|_2^2\)
contains shift energy plus finite-sample error energy in expectation.
Subtracting the estimated error energy
\(\widehat{\nu}_M\)
and normalizing by the observed energy gives the plug-in reliability factor
\begin{equation}
    \beta
    =
    \left[
    1-
    \frac{\widehat{\nu}_M}
    {\|\widehat{\mathbf b}_T\|_2^2}
    \right]_0^1,
    \label{eq:adaptive_shrinkage}
\end{equation}
where
\([x]_0^1=\min\{1,\max\{0,x\}\}\).
We set \(\beta=0\) when
\(\|\widehat{\mathbf b}_T\|_2^2=0\).
Thus, \(\beta\) approaches zero when the estimated shift is dominated by sampling uncertainty and approaches one when most of its observed energy is supported beyond that uncertainty.

SA-LBC therefore applies the shrunk shift to each subsequent target TEST observation \(q\) (Fig.~\ref{fig:framework}(d)),
\begin{equation}
    \widetilde{\mathbf z}_{T,q}
    =
    \mathbf z_{T,q}
    -
    \beta
    \widehat{\mathbf b}_T.
    \label{eq:test_correction}
\end{equation}
Calibration is performed before \(\ell_2\) normalization; normalization is applied only for cosine-similarity retrieval.

\subsection{Finite-Population Analysis of First-Order Estimation}

We isolate the effect of stimulus anchoring from reliability-aware shrinkage by analyzing the raw anchored and global-mean estimators.
We condition on a fixed target subject and its full \(N_T\)-stimulus calibration population; the only randomness is which \(M\) stimuli are sampled uniformly without replacement.

We define the full-population first-order shift as
\begin{equation}
    \mathbf b_T^{\mathrm{full}}
    =
    \frac{1}{N_T}
    \sum_{i=1}^{N_T}
    \left(
    \mathbf z_{T,i}
    -
    \mathbf p_i
    \right).
    \label{eq:full_train_bias_reference}
\end{equation}
This quantity serves only as the analysis reference and is not available to the deployed calibration procedure.

To separate stimulus composition from residual variation, define the centered anchor deviation and centered matched residual as
\begin{equation}
    \mathbf a_i
    =
    \mathbf p_i
    -
    \boldsymbol{\mu}_P,
    \qquad
    \boldsymbol{\delta}_{T,i}
    =
    \left(
    \mathbf z_{T,i}
    -
    \mathbf p_i
    \right)
    -
    \mathbf b_T^{\mathrm{full}}.
    \label{eq:centered_anchor_residual}
\end{equation}
By construction,
\begin{equation}
    \sum_{i=1}^{N_T}\mathbf a_i=\mathbf 0,
    \qquad
    \sum_{i=1}^{N_T}\boldsymbol{\delta}_{T,i}=\mathbf 0.
\end{equation}

For a sampled calibration subset \(\mathcal C_M\), let
\begin{equation}
    \bar{\mathbf a}_M
    =
    \frac{1}{M}
    \sum_{i\in\mathcal C_M}
    \mathbf a_i,
    \qquad
    \bar{\boldsymbol{\delta}}_M
    =
    \frac{1}{M}
    \sum_{i\in\mathcal C_M}
    \boldsymbol{\delta}_{T,i}.
\end{equation}
The estimator errors then decompose exactly as
\begin{align}
    \widehat{\mathbf b}_T^{\mathrm{SA}}
    -
    \mathbf b_T^{\mathrm{full}}
    &=
    \bar{\boldsymbol{\delta}}_M,
    \\
    \widehat{\mathbf b}_T^{\mathrm{GM}}
    -
    \mathbf b_T^{\mathrm{full}}
    &=
    \bar{\boldsymbol{\delta}}_M
    +
    \bar{\mathbf a}_M.
    \label{eq:estimator_error_decomposition}
\end{align}
Hence,
\begin{equation}
    \widehat{\mathbf b}_T^{\mathrm{GM}}
    -
    \widehat{\mathbf b}_T^{\mathrm{SA}}
    =
    \bar{\mathbf a}_M.
    \label{eq:exact_decomposition}
\end{equation}
Under uniform sampling, the composition term
\(\bar{\mathbf a}_M=\bar{\mathbf p}_M-\boldsymbol{\mu}_P\)
has zero expectation.
The unshrunk anchored estimator therefore has the classical
control-variate form with a unit coefficient~\cite{glynn2002controlvariates}.
This preserves unbiasedness but does not guarantee lower MSE:
the removed composition term \(\bar{\mathbf a}_M\) can covary
with the remaining residual error \(\bar{\boldsymbol{\delta}}_M\).

Their finite-population second-order structure is characterized by
\begin{align}
    \mathbf\Sigma_{\delta}
    &=
    \frac{1}{N_T-1}
    \sum_{i=1}^{N_T}
    \boldsymbol{\delta}_{T,i}
    \boldsymbol{\delta}_{T,i}^{\top},
    \\
    \mathbf\Sigma_{P}
    &=
    \frac{1}{N_T-1}
    \sum_{i=1}^{N_T}
    \mathbf a_i
    \mathbf a_i^{\top},
    \\
    \mathbf\Sigma_{\delta P}
    &=
    \frac{1}{N_T-1}
    \sum_{i=1}^{N_T}
    \boldsymbol{\delta}_{T,i}
    \mathbf a_i^{\top}.
    \label{eq:finite_population_covariances}
\end{align}
Here,
\(\mathbf\Sigma_{\delta}\)
measures residual variation,
\(\mathbf\Sigma_P\)
measures stimulus-anchor variation, and
\(\mathbf\Sigma_{\delta P}\)
captures their coupling.

\begin{theorem}[Exact finite-population MSE]
\label{thm:finite_population_mse}

If \(\mathcal C_M\) is sampled uniformly without replacement, both estimators are unbiased for
\(\mathbf b_T^{\mathrm{full}}\),
\begin{equation}
    \mathbb E
    \left[
    \widehat{\mathbf b}_{T}^{\mathrm{SA}}
    \right]
    =
    \mathbb E
    \left[
    \widehat{\mathbf b}_{T}^{\mathrm{GM}}
    \right]
    =
    \mathbf b_T^{\mathrm{full}}.
\end{equation}
The anchored estimator has mean-squared error
\begin{equation}
    \mathbb E
    \left[
    \left\|
    \widehat{\mathbf b}_{T}^{\mathrm{SA}}
    -
    \mathbf b_T^{\mathrm{full}}
    \right\|_2^2
    \right]
    =
    \kappa_M
    \operatorname{tr}
    \left(
    \mathbf\Sigma_{\delta}
    \right),
    \label{eq:mse_sa_general}
\end{equation}
where
\begin{equation}
    \kappa_M
    =
    \frac{1}{M}
    -
    \frac{1}{N_T}.
    \label{eq:finite_population_factor}
\end{equation}
The corresponding global-mean MSE is
\begin{equation}
\begin{aligned}
    &
    \mathbb E
    \left[
    \left\|
    \widehat{\mathbf b}_{T}^{\mathrm{GM}}
    -
    \mathbf b_T^{\mathrm{full}}
    \right\|_2^2
    \right]
    \\
    &\quad=
    \kappa_M
    \left[
    \operatorname{tr}
    \left(
    \mathbf\Sigma_{\delta}
    \right)
    +
    \operatorname{tr}
    \left(
    \mathbf\Sigma_P
    \right)
    +
    2
    \operatorname{tr}
    \left(
    \mathbf\Sigma_{\delta P}
    \right)
    \right].
    \label{eq:mse_gm_general}
\end{aligned}
\end{equation}

\end{theorem}

The proof uses the finite-population sample-mean covariance identity~\cite{cochran1977sampling} and is given in Supplementary Section~S2.

Subtracting the two MSEs gives
\begin{equation}
\begin{aligned}
    \Delta_{\mathrm{MSE}}
    &=
    \operatorname{MSE}
    \left(
    \widehat{\mathbf b}_{T}^{\mathrm{GM}}
    \right)
    -
    \operatorname{MSE}
    \left(
    \widehat{\mathbf b}_{T}^{\mathrm{SA}}
    \right)
    \\
    &=
    \kappa_M
    \left[
    \operatorname{tr}
    \left(
    \mathbf\Sigma_P
    \right)
    +
    2
    \operatorname{tr}
    \left(
    \mathbf\Sigma_{\delta P}
    \right)
    \right].
    \label{eq:generalized_mse_gap}
\end{aligned}
\end{equation}

\begin{corollary}[Condition for lower anchored-estimator MSE]
\label{cor:anchoring_advantage}

For \(M<N_T\),
\begin{equation}
\begin{gathered}
\operatorname{MSE}\!\left(\widehat{\mathbf b}_{T}^{\mathrm{SA}}\right)
<\operatorname{MSE}\!\left(\widehat{\mathbf b}_{T}^{\mathrm{GM}}\right)
\\\Longleftrightarrow\quad
\operatorname{tr}(\mathbf\Sigma_{\delta P})
> -\tfrac12\operatorname{tr}(\mathbf\Sigma_P).
\end{gathered}
\label{eq:generalized_advantage_condition}
\end{equation}

\end{corollary}

Stimulus anchoring therefore has lower MSE only when residual--anchor coupling is not sufficiently negative to offset the anchor-variation term.
In the special case
\(\mathbf\Sigma_{\delta P}=\mathbf 0\),
anchoring has lower MSE whenever
\(\operatorname{tr}(\mathbf\Sigma_P)>0\).
No residual--anchor orthogonality assumption is required by the general result.

The estimator difference in Eq.~\eqref{eq:exact_decomposition} also gives
\begin{equation}
    \mathbb E
    \left[
    \left\|
    \widehat{\mathbf b}_{T}^{\mathrm{GM}}
    -
    \widehat{\mathbf b}_{T}^{\mathrm{SA}}
    \right\|_2^2
    \right]
    =
    \kappa_M
    \operatorname{tr}
    \left(
    \mathbf\Sigma_P
    \right).
    \label{eq:composition_energy_general}
\end{equation}
This stimulus-composition energy decreases with calibration budget and vanishes at \(M=N_T\).
When the condition in
Corollary~\ref{cor:anchoring_advantage}
holds, the positive MSE gap is scaled by the same factor
\(\kappa_M\)
and therefore shrinks as \(M\) approaches \(N_T\).

The cross-covariance term
\(\mathbf\Sigma_{\delta P}\)
also allows residual variation to remain coupled to stimulus geometry, so the theorem does not require purely additive cross-subject mismatch.
For illustration, consider
\begin{equation}
\begin{aligned}
\mathbf z_{T,i}
&=\boldsymbol{\mu}_P+\mathbf b_T^{\mathrm{full}}
  +\mathbf Q_T\mathbf a_i+\boldsymbol{\epsilon}_i,\\
\mathbf Q_T^{\top}\mathbf Q_T&=\mathbf I,\qquad
\sum_{i=1}^{N_T}\boldsymbol{\epsilon}_i=\mathbf 0.
\end{aligned}
\label{eq:illustrative_rotation_model}
\end{equation}
Since
\(\mathbf p_i=\boldsymbol{\mu}_P+\mathbf a_i\),
the matched residual becomes
\begin{equation}
    \mathbf z_{T,i}-\mathbf p_i
    =
    \mathbf b_T^{\mathrm{full}}
    +
    (\mathbf Q_T-\mathbf I)\mathbf a_i
    +
    \boldsymbol{\epsilon}_i.
\end{equation}
Thus, reorientation can induce residual--anchor coupling through
\((\mathbf Q_T-\mathbf I)\mathbf a_i\).
This model is illustrative only and is not assumed by the theorem.
The finite-population analysis characterizes estimation within the calibration pool; transfer to concept-disjoint TEST stimuli is assessed empirically.

\subsection{Calibration Budget and Orientation Estimation}\label{sec:orientation}

The finite-population analysis allows higher-order mismatch, but does not determine how much orientation information can be estimated from the same few-shot calibration set.
Let
\(\mathcal C_M=\{i_1,\ldots,i_M\}\).
To isolate centered geometry, we form the matched calibration matrices
\begin{equation}
    \mathbf X_M
    =
    \begin{bmatrix}
    (\mathbf z_{T,i_1}-\bar{\mathbf z}_{T,M})^{\top}\\
    \vdots\\
    (\mathbf z_{T,i_M}-\bar{\mathbf z}_{T,M})^{\top}
    \end{bmatrix},
    \qquad
    \mathbf Y_M
    =
    \begin{bmatrix}
    (\mathbf p_{i_1}-\bar{\mathbf p}_{M})^{\top}\\
    \vdots\\
    (\mathbf p_{i_M}-\bar{\mathbf p}_{M})^{\top}
    \end{bmatrix}.
\end{equation}
Because the centered rows sum to zero,
\begin{equation}
    \operatorname{rank}(\mathbf X_M),
    \operatorname{rank}(\mathbf Y_M)
    \le M-1,
\end{equation}
and therefore
\begin{equation}
    \operatorname{rank}
    \left(
    \mathbf X_M^{\top}\mathbf Y_M
    \right)
    \le M-1.
    \label{eq:orientation_rank_bound}
\end{equation}
The cross-covariance therefore contains at most \(M-1\) nonzero singular modes.
The union of the centered target and anchor spans can have dimension up to \(2(M-1)\); the rank bound does not restrict an orthogonal correction to only \(M-1\) ambient directions.
In our main setting, \(d=768\) and
\(M\in\{2,4,8,16\}\),
giving rank bounds of \(1\), \(3\), \(7\), and \(15\), respectively.
This rank bound neither rules out reorientation nor implies that translation must perform better; it only limits the orientation structure directly supported by the observed pairs.

To evaluate orientation under the same information budget, we estimate an identity-regularized orthogonal map
\begin{equation}
\begin{aligned}
\widehat{\mathbf R}
=\underset{\mathbf R^\top\mathbf R=\mathbf I}{\arg\min}\;
\Bigl\{&\|\mathbf X_M\mathbf R-\mathbf Y_M\|_F^2\\
&+\lambda_R\|\mathbf R-\mathbf I\|_F^2\Bigr\}.
\end{aligned}
\label{eq:orthogonal_control}
\end{equation}
where
\begin{equation}
    \lambda_R
    =
    \rho
    \left\|
    \mathbf X_M^\top\mathbf Y_M
    \right\|_2.
\end{equation}
Here, \(\rho\ge0\) controls identity regularization and
\(\|\cdot\|_2\) denotes the spectral norm.
By the orthogonal Procrustes solution~\cite{schonemann1966procrustes}, if
\begin{equation}
    \mathbf X_M^\top\mathbf Y_M
    +
    \lambda_R\mathbf I
    =
    \mathbf U\mathbf\Sigma\mathbf V^\top,
\end{equation}
then
\begin{equation}
    \widehat{\mathbf R}
    =
    \mathbf U\mathbf V^\top.
\end{equation}

Because calibration observations are stacked as rows, the corresponding transform on column-vector embeddings is
\(\widehat{\mathbf R}^{\top}\).
The orientation-only correction therefore preserves the target calibration centroid,
\begin{equation}
    \widetilde{\mathbf z}_{T,q}^{R}
    =
    \bar{\mathbf z}_{T,M}
    +
    \widehat{\mathbf R}^{\top}
    \left(
    \mathbf z_{T,q}
    -
    \bar{\mathbf z}_{T,M}
    \right).
    \label{eq:r_only_control}
\end{equation}
The combined correction shifts the centroid by SA-LBC while applying the same orientation map to centered structure,
\begin{equation}
\begin{aligned}
\widetilde{\mathbf z}_{T,q}^{T+R}
={}&\bar{\mathbf z}_{T,M}-\beta\widehat{\mathbf b}_T\\
&+\widehat{\mathbf R}^{\top}
\left(\mathbf z_{T,q}-\bar{\mathbf z}_{T,M}\right).
\end{aligned}
\label{eq:tr_control}
\end{equation}
These controls use the same frozen representation, matched observations, and calibration budget, holding target information access fixed while varying the geometric form of the correction.

\section{Experiments}
\subsection{Experimental Setup}
\paragraph{Data and information access}
THINGS-EEG2~\cite{Things-EEG2} is the primary benchmark, with 10 subjects evaluated by leave-one-subject-out (LOSO) training on the other nine.
The 200-image TEST gallery is concept-disjoint from TRAIN.
EEG is whitened within each participant and session using that participant's retained TRAIN epochs, not TEST covariance; the encoder receives 17 channels and 250 samples.
TRAIN and TEST observations average four and 80 repetitions per stimulus, respectively.
Thus, $M$ counts stimulus-level calibration averages after participant-specific preprocessing, not total target recordings or single-trial inputs.
For each of 200 random permutations of target TRAIN, prefixes of lengths $M\in\{2,4,8,16\}$ define nested calibration sets shared across methods and seeds.
Stimulus identities index source neural anchors; calibration uses no target image pixels, image embeddings, semantic labels, TEST statistics, retrieval feedback, or representation updates.
Alljoined-1.6M~\cite{xu2025alljoined16mmilliontrialeegimagedataset} provides a 20-subject EEG cohort; THINGS-MEG~\cite{Things-data} provides four subjects and two predefined variants for descriptive boundary analysis.
Preprocessing, valid-repeat averaging, and dataset-specific training protocols are detailed in Supplementary Section~S1; external results are in Section~S6.

\paragraph{Frozen representations}
The primary diagnostic configuration uses our attention-based EEG
encoder (denoted Attention) with a frozen CLIP ViT-B/32
backbone~\cite{CLIP}.
Source-trained projections map both modalities into a
768-dimensional retrieval space; visual prompts are disabled.
Source-support, shift-fidelity, and directional diagnostics use this configuration.
Source-balanced multi-positive contrastive training uses source-TRAIN concept-disjoint checkpoint selection only.
We compare Attention, EEGProject, and TSConv with CLIP ViT-B/32; Attention with CLIP RN101 and DINOv2 ViT-B/14~\cite{oquab2024dinov2}; and TSConv with DINOv2.
These are six configurations, counting Attention--CLIP ViT-B/32 once, not a complete Cartesian product.
Each configuration independently fits its EEG encoder and projections with three seeds, while keeping its pretrained visual backbone frozen; fixing an architecture does not fix its learned weights.
All six use the same source-only training/selection protocol and paired draws without architecture-specific tuning.
Architecture and optimizer details are in Supplementary Tables~S1 and S11.

\paragraph{Controls}
We compare no adaptation, raw global-mean matching, adaptive global mean, bank-shuffled anchoring, raw anchoring, and SA-LBC.
Adaptive global mean shares SA-LBC's source-derived noise term but computes its own coefficient:
\begin{equation}
\begin{aligned}
\beta_{\mathrm{GM}}
&=\left[1-\frac{\widehat{\nu}_M}
{\|\widehat{\mathbf b}_{T}^{\mathrm{GM}}\|_2^2}\right]_0^1,\\
\widetilde{\mathbf z}_{T,q}^{\mathrm{GM}}
&=\mathbf z_{T,q}-\beta_{\mathrm{GM}}\widehat{\mathbf b}_{T}^{\mathrm{GM}}.
\end{aligned}
\label{eq:adaptive_global_control}
\end{equation}
The coefficient is zero for a zero-norm estimate.
This comparison changes reference construction and its induced coefficient, not the reference alone at fixed strength.
The bank-shuffled control instead fixes the matched-condition coefficient and replaces anchors through four derangements of the full source bank per draw.
It generally changes the selected anchor set, testing stimulus-composition matching rather than pairing order within a fixed set.
Raw anchoring fixes $\beta=1$.
The orientation controls in Section~\ref{sec:orientation} use the same matched data, with primary $\rho=0.1$ and a fixed nine-value sensitivity grid; no TEST-based deployment selection is performed.
They isolate regularized orthogonal calibration, not the complete SCORE method~\cite{cuiSCORESubjectCoordinate2026}.
Norm-matched random and sign-flipped directions are additional controls for Attention--CLIP ViT-B/32; full definitions are in Supplementary Section~S7.7.

\paragraph{Statistics}
MRR is the primary endpoint; Top-1 and Top-5 are secondary metrics.
Draws and training seeds are averaged within each held-out subject before reporting subject-level means, SDs, and inference.
THINGS-EEG2 contrasts use exact two-sided paired sign-flip tests and 20,000 subject-bootstrap resamples~\cite{efron1979bootstrap} for 95\% intervals.
Holm adjustment~\cite{holm1979sequential} covers four budgets separately within each contrast and configuration, not all configurations or contrasts jointly.
The $4\times9$ regularization grids and joint-configuration contrasts are descriptive.
Alljoined uses 20 subjects, 50,000 bootstrap resamples, exact sign flips, and the explicitly supplemental within-contrast Holm-4 analysis; MEG remains descriptive over four subjects.
Full statistical families and convergence details are in Supplementary Table~S3 and Fig.~S1.

\subsection{Calibration Utility and Source-Reference Matching}
Table~\ref{tab:things_main} reports all budgets and controls for Attention--CLIP ViT-B/32.
SA-LBC raises MRR from 0.1867 to 0.2013 at $M=2$ and 0.2336 at $M=16$.
At $M=16$, Top-1 increases from 8.63\% to 12.10\% and Top-5 from 26.20\% to 33.11\%.
MRR and Top-5 improve in all 10 subjects at every budget; Top-1 improves in 9/10 subjects at $M=2,4$ and 10/10 at $M=8,16$.
Each SA-LBC--baseline MRR contrast has $p_{\mathrm{Holm}}=0.0078$.
These are configuration-specific gains, not representation-invariant benefits.

\begin{table*}[t]
\centering
\caption{Attention--CLIP ViT-B/32 retrieval after participant-specific preprocessing. MRR is the primary endpoint; values are mean $\pm$ SD over 10 subjects after averaging 200 paired draws and three seeds within subject. Both adaptive estimators share the noise term but compute their own coefficient; raw anchoring uses $\beta=1$. Bold marks the highest rounded mean within each calibration budget. Complete Top-1 and Top-5 results are in Supplementary Table~S12.}
\label{tab:things_main}
\small
\setlength{\tabcolsep}{6pt}
\begin{tabular}{lcccc}
\toprule
Method & $M=2$ & $M=4$ & $M=8$ & $M=16$ \\
\midrule
No adaptation & 0.1867 $\pm$ 0.0495 & 0.1867 $\pm$ 0.0495 & 0.1867 $\pm$ 0.0495 & 0.1867 $\pm$ 0.0495 \\
Global mean & 0.1673 $\pm$ 0.0265 & 0.1994 $\pm$ 0.0338 & 0.2213 $\pm$ 0.0412 & 0.2347 $\pm$ 0.0459 \\
Adaptive global mean & 0.1998 $\pm$ 0.0459 & 0.2107 $\pm$ 0.0460 & 0.2219 $\pm$ 0.0475 & 0.2318 $\pm$ 0.0489 \\
Shuffled anchor & 0.1953 $\pm$ 0.0456 & 0.2056 $\pm$ 0.0451 & 0.2166 $\pm$ 0.0456 & 0.2275 $\pm$ 0.0470 \\
Raw anchored ($\beta=1$) & 0.1822 $\pm$ 0.0341 & 0.2090 $\pm$ 0.0401 & \textbf{0.2271 $\pm$ 0.0450} & \textbf{0.2381 $\pm$ 0.0481} \\
SA-LBC & \textbf{0.2013 $\pm$ 0.0464} & \textbf{0.2120 $\pm$ 0.0476} & 0.2239 $\pm$ 0.0495 & 0.2336 $\pm$ 0.0504 \\
\bottomrule
\end{tabular}
\end{table*}

\begin{figure*}[t]
\centering
\subfloat[]{\includegraphics[width=0.31\textwidth]{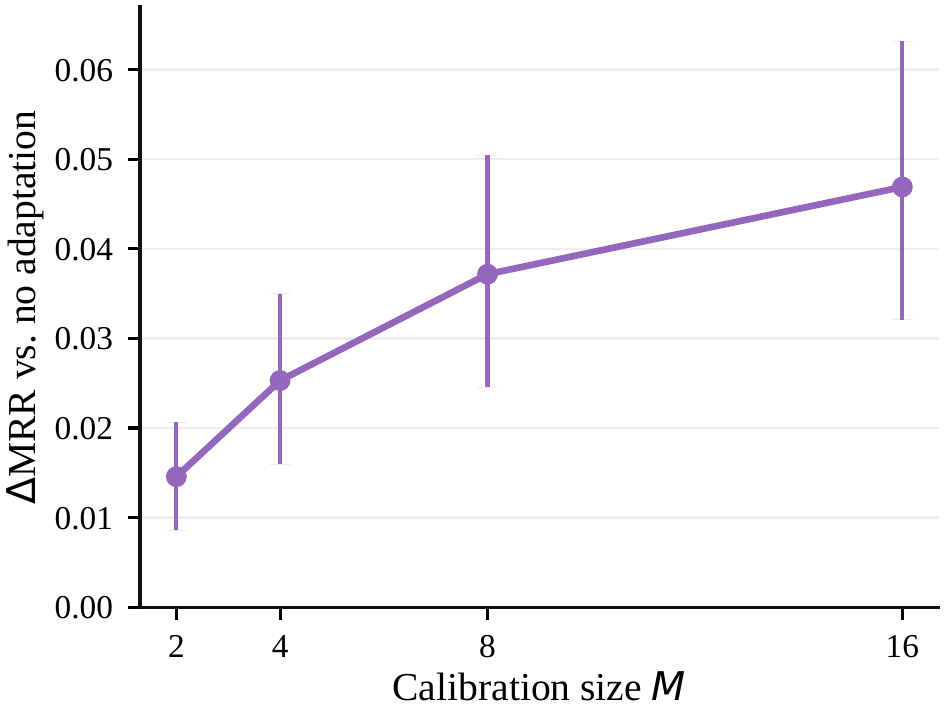}}\hfill
\subfloat[]{\includegraphics[width=0.31\textwidth]{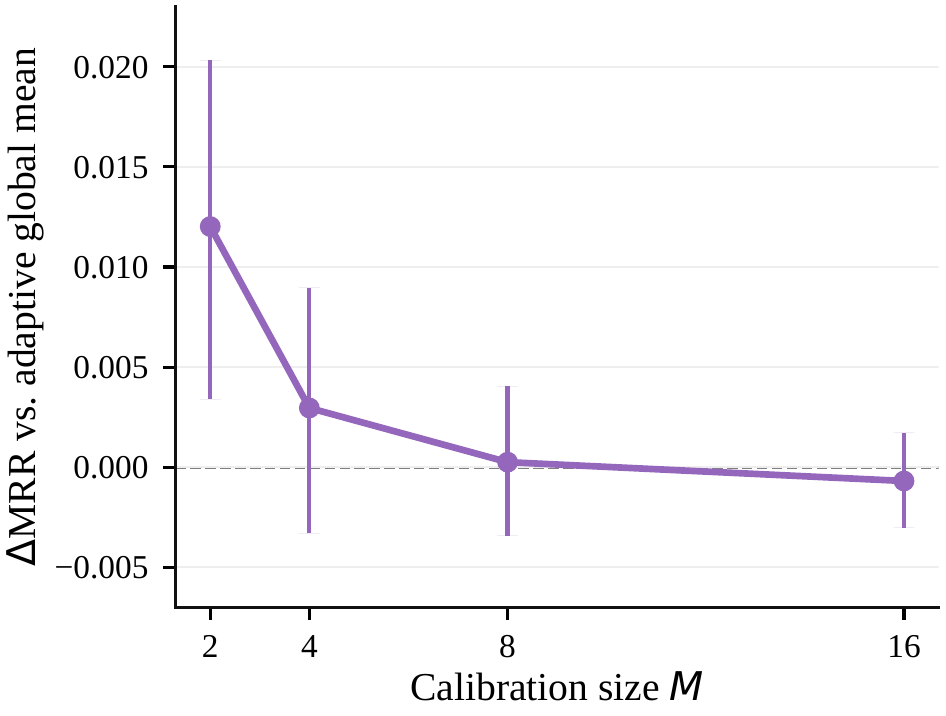}}\hfill
\subfloat[]{\includegraphics[width=0.31\textwidth]{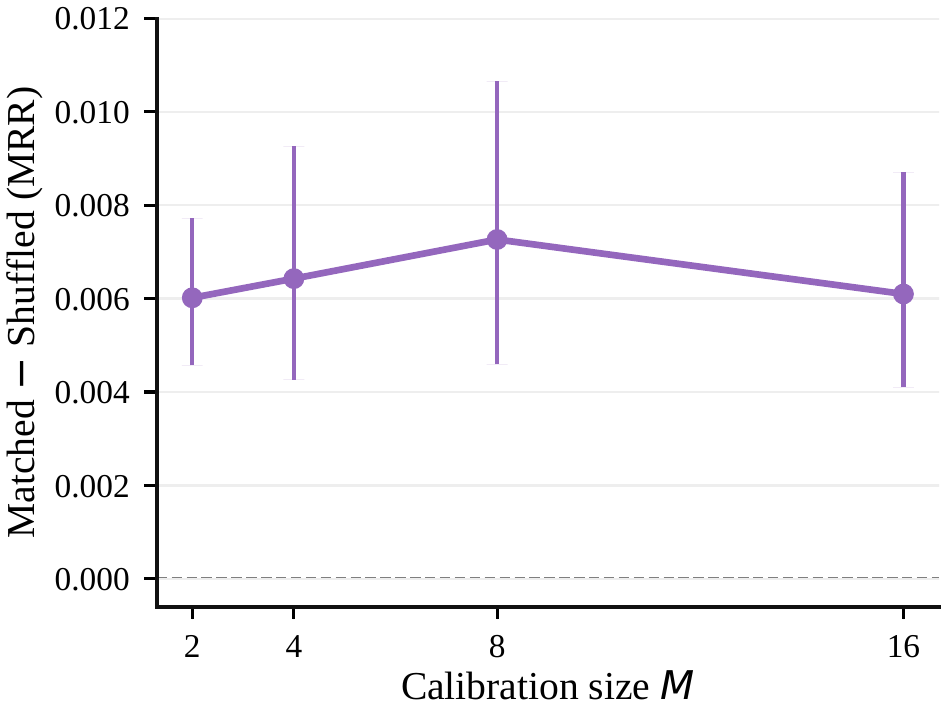}}
\caption{Paired MRR effects for Attention--CLIP ViT-B/32: SA-LBC minus (a) no adaptation, (b) adaptive global mean, and (c) bank-shuffled anchoring. Error bars are 95\% subject-bootstrap intervals after within-subject draw/seed averaging. The controls test net utility, the adaptive-reference increment, and reference-composition matching, respectively; their effects are not interchangeable.}
\label{fig:paired_effects}
\end{figure*}

Against adaptive global mean, the paired MRR gains are 0.00151, 0.00133, 0.00204, and 0.00183 at $M=2,4,8,16$, respectively, with $p_{\mathrm{Holm}}=0.0117,0.0273,0.0078,0.0078$.
The increment is small and not monotonically decreasing with budget.
Matched references exceed bank-shuffled references by 0.0060--0.0073 MRR, with positive differences in all subjects and $p_{\mathrm{Holm}}=0.0078$ at each budget (Fig.~\ref{fig:paired_effects}).
Omitting matching and supplying a mismatched reference are different controls; neither replaces the no-adaptation comparison.
The finite-population decrease in composition energy does not impose a corresponding MRR trend after estimator-specific shrinkage and cosine normalization.

Increasing source support $K$ improves group-average calibration gains under the same rule, but both the noise term and coefficient are recomputed as $K$ changes.
This is not an isolated test of anchor averaging at fixed strength (Supplementary Table~S6 and Section~S7.3).

\subsection{Verification of First-Order Estimation}
For each configuration, subject, and seed, analytic MSE uses the full target-TRAIN finite population, and empirical MSE uses 200 calibration draws of the unshrunk estimators.
The full-TRAIN residual mean serves only as the retrospective analysis reference and is not used to estimate or select the deployed correction.
Across 120 seed--subject--budget conditions per configuration, mean relative discrepancies are 0.90--1.61\% for anchoring and 1.05--1.58\% for global-mean estimation; ordering agreement is 90.0--100.0\% (Supplementary Table~S33).
This is agreement between analytic and empirical error rankings, not the fraction in which anchoring wins.
The audit verifies subset-sampling calculations, not a retrieval ordering after shrinkage, normalization, and transfer to concept-disjoint TEST stimuli.

\subsection{Geometry and Correction Strength}
We compare SA-LBC translation ($T$), centroid-preserving regularized orientation ($R$), and their combination ($T+R$) using the same frozen representations and matched calibration observations.
The centered cross-covariance respects rank bounds of 1, 3, 7, and 15 for $M=2,4,8,16$; union-span dimensions do not exceed 2, 6, 14, and 30 in the 768-dimensional space.
These constraints do not determine retrieval ordering.
At $\rho=0.1$, $T>R$ and $T+R<T$ across all six configurations and four budgets, with the same direction in all subjects and $p_{\mathrm{Holm}}=0.0078$ within each contrast/configuration.
At $\rho=1$, $R$ instead exceeds $T$ at every budget for both TSConv configurations in group-average MRR.
At the same $\rho=1$ and $M=16$, $T+R$ exceeds $T$ in all six configurations by 0.00377--0.00790 (Table~\ref{tab:geometry_order}).
These grid results are descriptive, not TEST-selected deployment settings; the primary disadvantage does not make orientation intrinsically harmful.

\begin{table*}[t]
\centering
\caption{Regularization-dependent geometric ordering at $M=16$.
Entries are group-average MRR from the same cache within each configuration.
Comparisons of correction types are made within each row.
$\rho=0.1$ is the primary setting; $\rho=1$ is a fixed member of the sensitivity grid, not a TEST-selected deployment parameter.
The complete grid is reported in the Supplementary Material.}
\label{tab:geometry_order}
\small
\setlength{\tabcolsep}{6pt}
\begin{tabular}{llccccc}
\toprule
EEG encoder & Visual backbone & $T$ & $R_{0.1}$ & $(T+R)_{0.1}$ & $R_{1}$ & $(T+R)_{1}$ \\
\midrule
Attention & CLIP ViT-B/32 & 0.23364 & 0.16749 & 0.20697 & 0.19326 & 0.24114 \\
EEGProject & CLIP ViT-B/32 & 0.25908 & 0.17956 & 0.21624 & 0.22060 & 0.26695 \\
TSConv & CLIP ViT-B/32 & 0.21047 & 0.18012 & 0.18171 & 0.21503 & 0.21582 \\
Attention & CLIP RN101 & 0.21311 & 0.15846 & 0.19034 & 0.18487 & 0.22101 \\
Attention & DINOv2 ViT-B/14 & 0.18318 & 0.13609 & 0.16186 & 0.15945 & 0.18855 \\
TSConv & DINOv2 ViT-B/14 & 0.18264 & 0.15619 & 0.15840 & 0.18479 & 0.18641 \\
\bottomrule
\end{tabular}
\end{table*}

Even within translation, correction strength remains a separate question.
Shrinkage improves on raw anchoring in all six configurations at $M=2$ ($p_{\mathrm{Holm}}<0.05$ within each configuration).
For Attention--CLIP ViT-B/32, MRR is 0.2013 versus 0.1822 ($\Delta=+0.0191$, $p_{\mathrm{Holm}}=0.0469$); raw anchoring falls below no adaptation (0.1867).
At $M=16$, raw anchoring instead reaches 0.2381 versus 0.2336 ($\Delta=-0.0045$, $p_{\mathrm{Holm}}=0.0391$).
Both TSConv configurations still benefit from shrinkage at $M=16$, by 0.00567 with CLIP and 0.00376 with DINOv2 (both $p_{\mathrm{Holm}}=0.0078$).
Thus, sample count alone does not determine a common crossover, and improving on raw correction does not guarantee improving on no adaptation.

Diagnostics for Attention--CLIP ViT-B/32 show increasing fidelity to the within-model full-TRAIN reference as $M$ grows.
Norm-matched random directions remain near no adaptation, whereas sign reversal degrades retrieval; SA-LBC exceeds both controls in all subjects and budgets ($p_{\mathrm{Holm}}=0.0078$ within each family).
Full magnitudes, fidelity, and direction results are retained in Supplementary Section~S7.4 and Tables~S8--S10.

\subsection{Representation and External Boundaries}
Table~\ref{tab:representation_dependence} compares each configuration
with its own same-cache baseline.
With CLIP ViT-B/32, EEGProject achieves higher group-average MRR
than Attention before calibration (0.21393 versus 0.18675)
and after SA-LBC at $M=16$ (0.25908 versus 0.23364).
Its larger final MRR does not reflect a larger paired calibration
gain (+0.04515 versus +0.04690).
SA-LBC improves Attention across all three evaluated visual backbones
and EEGProject with CLIP across all budgets.
TSConv--CLIP instead has a negative $M=2$ effect in all subjects ($\Delta\mathrm{MRR}=-0.00330$, $p_{\mathrm{Holm}}=0.0078$), and only +0.00100 at $M=16$ ($p_{\mathrm{Holm}}=0.6172$).
TSConv--DINOv2 effects are -0.00236 and +0.00143 at those budgets, neither significant after correction ($p_{\mathrm{Holm}}=0.0547,0.2754$).
The configurations are separately source-trained; these comparisons do not isolate architecture causality or establish an encoder--backbone interaction.
Matched references still exceed shuffled references in every configuration and budget, whereas their advantage over adaptive global mean is not significant at any budget for TSConv--DINOv2.
Reference matching is therefore not a guarantee of useful calibration.

\begin{table*}[t]
\centering
\caption{Calibration utility across six source-trained neural--visual configurations.
Baseline MRR and paired changes use the same re-encoded cache within each row; Attention--CLIP ViT-B/32 is counted once.
$\Delta$ is SA-LBC minus no adaptation.
$p_{\mathrm{H}}$ is Holm-adjusted across $M\in\{2,4,8,16\}$ separately within each configuration and contrast, not across the six rows. These tests do not compare encoders.
The full four-budget results are provided in the Supplementary Material.}
\label{tab:representation_dependence}
\small
\setlength{\tabcolsep}{4pt}
\begin{tabular}{llccccc}
\toprule
EEG encoder & Visual backbone & \thead{Baseline\\MRR} & \thead{$\Delta$MRR\\($M=2$)} & $p_{\mathrm{H}}$ & \thead{$\Delta$MRR\\($M=16$)} & $p_{\mathrm{H}}$ \\
\midrule
Attention & CLIP ViT-B/32 & 0.18675 & +0.01458 & 0.0078 & +0.04690 & 0.0078 \\
EEGProject & CLIP ViT-B/32 & 0.21393 & +0.01224 & 0.0234 & +0.04515 & 0.0078 \\
TSConv & CLIP ViT-B/32 & 0.20947 & -0.00330 & 0.0078 & +0.00100 & 0.6172 \\
Attention & CLIP RN101 & 0.18022 & +0.00944 & 0.0215 & +0.03290 & 0.0078 \\
Attention & DINOv2 ViT-B/14 & 0.15587 & +0.00800 & 0.0078 & +0.02730 & 0.0078 \\
TSConv & DINOv2 ViT-B/14 & 0.18122 & -0.00236 & 0.0547 & +0.00143 & 0.2754 \\
\bottomrule
\end{tabular}
\end{table*}

External results use three seeds and 200 paired draws (Supplementary Section~S6 and Fig.~S6).
Alljoined MRR changes are -0.00048, +0.00044, +0.00120, and +0.00228 as $M$ increases.
At $M=16$, 16/20 subjects improve; the unadjusted 95\% interval is [0.00047,0.00418] and exact $p=0.0281$, but supplemental Holm-4 gives $p=0.1124$.
No SA-LBC--baseline MRR contrast passes that correction.
Matched references do exceed shuffled references by 0.00025--0.00032 under their own Holm-4 correction, which does not establish a net calibration benefit.
MEG gains at $M=16$ are +0.00265 and +0.01009 for the baseline and full variants.
With four subjects and multiple components changed together, this is descriptive evidence, not a causal or modality-general result.

Finally, the dedicated SATTC comparison uses the same preprocessed inputs but different latent-calibration information: 200 unlabeled TEST queries versus SA-LBC's matched pre-test observations.
It is a regime comparison, not a matched-information algorithmic ranking (Supplementary Section~S7.6 and Table~S35).

\section{Discussion}
Our results separate the information needed to estimate a cross-subject correction from the conditions under which that correction improves retrieval.
Stimulus matching controls the source reference, shrinkage controls the applied strength, and geometric regularization controls flexibility.
These choices remain constrained by the calibration budget, but their functional value also depends on the frozen representation.
The resulting view is not that one correction is universally preferable, but that calibration must be evaluated under its information and representation conditions.

Stimulus anchoring has a specific estimation role: it replaces a fixed population centroid with the source reference for the sampled target stimuli.
This removes the explicit composition term in the global-mean estimator, but does not guarantee lower MSE because residual--anchor cross-covariance can offset the removed variation.
The finite-population analysis states that boundary exactly, and the sampling audits verify its numerical predictions across the six configurations.
Retrieval supplies a separate test: matched anchors consistently outperform shuffled anchors, whereas their advantage over adaptive global mean is smaller and not significant in every configuration.
The two controls cannot be conflated, because omitting correspondence differs from imposing an incorrect correspondence.
Nor does decreasing composition energy imply a decreasing MRR gap after shrinkage and normalization.
Relative to prior uses of shared stimuli for cross-subject alignment~\cite{haxbyCommonHighDimensionalModel2011,ferranteTheirEyesMultisubject2024a,daiMindAlignerExplicitBrain2025,liuMindAdapterFewShotParameterEfficient2026}, the contribution is therefore an explicit account of how reference construction changes few-shot estimation error, not the first use of correspondence itself.

Reference construction does not determine which geometric transformation the same observations can support.
At the primary regularization setting, translation outperforms standalone orientation in all evaluated configurations and budgets.
However, stronger identity regularization reverses this ordering for the TSConv configurations and allows orientation to complement translation.
The low rank of the centered cross-covariance limits data-supported geometry, but is not a proof of translation's superiority.
These results qualify rather than contradict evidence that richer mappings can support cross-subject alignment~\cite{rodriguesRiemannianProcrustesAnalysis2019a,cuiSCORESubjectCoordinate2026}.
They show why geometric class and regularization must be considered together; the sensitivity grid does not provide a TEST-free rule for choosing either.

Even for a fixed transformation class, correction strength remains a separate estimation decision.
At the smallest budget, shrinkage improves on the raw anchored correction in all six configurations.
Its larger-budget behavior is not uniform: full-strength correction outperforms shrinkage for Attention--CLIP ViT-B/32 at \(M=16\), whereas both TSConv configurations still benefit from shrinkage.
Sample count alone therefore does not determine when attenuation becomes unnecessary.
These contrasts also distinguish limiting an unreliable correction from ensuring a net retrieval gain.
The crossover in this configuration is consistent with improving shift fidelity, but does not distinguish uncertainty misestimation from a mismatch between squared-error estimation and ranking-based utility.
The current coefficient remains a source-derived plug-in rule, not retrieval-optimal shrinkage.

The six-configuration results make this distinction concrete.
With CLIP ViT-B/32, EEGProject has higher group-average MRR than Attention
before and after calibration, but smaller mean calibration gains at
\(M=2\) and \(M=16\).
SA-LBC benefits Attention with each evaluated visual backbone and
EEGProject with CLIP, whereas TSConv yields small or negative calibration
effects with both CLIP and DINOv2.
Yet matched anchors remain preferable to shuffled anchors in those same TSConv configurations.
Correct correspondence can therefore improve a calibration procedure without making that procedure better than no calibration.
This is a configuration-dependent result, not evidence that neural architecture alone causes the difference: each configuration has separately learned source alignment, and visual changes also alter pretraining and preprocessing.
The external results extend the boundary rather than remove it.
Alljoined-1.6M shows smaller, heterogeneous gains, while the two THINGS-MEG variants provide only descriptive evidence from four subjects.
Neither establishes universal cross-dataset or cross-modality benefit, and the mechanism linking representation to calibration utility remains unresolved.

These findings must also be interpreted within the information available to each method.
Shared-space and functional-alignment approaches use shared responses or stimulus correspondence~\cite{haxbyCommonHighDimensionalModel2011,ferranteTheirEyesMultisubject2024a}, whereas SATTC and SCORE exploit unlabeled target-side structure at deployment~\cite{huangSATTCStructureAwareLabelFree,cuiSCORESubjectCoordinate2026}.
Conditional on the stated preprocessing, SA-LBC uses a frozen source-trained representation and a small set of matched observations to estimate the latent correction, without using future target TEST statistics or retrieval feedback.
The dedicated SATTC comparison is therefore a comparison of information regimes, not a matched-information test of algorithmic superiority.
Calibration methods should be distinguished by when target data become available, whether correspondence is supplied, and whether future TEST observations are jointly accessible.
Under this distinction, SA-LBC is stimulus-matched pre-test calibration, not unrestricted label-free adaptation.

Several limits remain.
Participant-specific TRAIN whitening uses target recordings beyond the latent-calibration subset. The present THINGS-EEG2 results therefore do not establish end-to-end \(M\)-observation adaptation. Testing that stronger claim would require budgeting all target-dependent preprocessing as well as calibration.
Known stimulus identities are required to retrieve the source anchors; extending beyond that regime requires estimating correspondence without target TEST information.
The geometric controls cover identity-regularized orthogonal maps, not arbitrary nonlinear transformations, and the six configurations do not constitute a full encoder--backbone factorial design.
Neither the sensitivity grid nor the shrinkage comparisons establish a source-only selector of geometric class and correction strength.
A stronger test would require such a selector, evaluated without target TEST feedback, as well as independent characterization of representation properties that predict calibration utility.
Broader validity requires larger external cohorts and tests across sessions, devices, and modalities.
Finally, the full-TRAIN residual mean is a retrospective finite-population reference, not ground-truth biological bias; fidelity to it does not establish recovery of a stable biological subject-specific state.

The practical implication is to distinguish estimating a correction, controlling its strength and flexibility, and establishing its utility relative to no calibration.
The available target evidence constrains each decision, but the appropriate decision also depends on the representation being corrected.
Cross-subject calibration should therefore ask not only what transformation can be expressed, but what correction the available evidence can support and whether applying it is useful.

\section{Conclusion}
\label{conclusion}
We study stimulus-matched latent calibration for cross-subject neural--image retrieval conditional on fixed preprocessing, and introduce SA-LBC to estimate a first-order shift with source-derived shrinkage.
The finite-population analysis gives exact errors for the unshrunk anchored and global-mean estimators and identifies when anchoring lowers MSE.
Across six source-trained configurations, correct stimulus correspondence contributes consistently, but net retrieval gains, the benefit of shrinkage, and geometric ordering are not representation invariant.
The primary translation advantage is conditional on regularization, and the TSConv results show that matching the source reference or attenuating
the correction need not yield a net retrieval benefit over no calibration.
These findings support evaluating reference construction, correction strength, and geometric flexibility against both the available target evidence and the frozen representation.

\section{Data and Code Availability}
The neural datasets used in this study are publicly available from
their original repositories: THINGS-EEG2~\cite{Things-EEG2},
Alljoined-1.6M~\cite{xu2025alljoined16mmilliontrialeegimagedataset}, and THINGS-MEG~\cite{Things-data}.
The source code, configuration files, and evaluation scripts required
to reproduce the reported experiments are publicly available at:
\url{https://github.com/Xizhang-Yao/SA-LBC}.

\section{Conflicts of Interest}
The authors declare that they have no conflicts of interest.
\section{Acknowledgments}
This work was supported by the National Natural Science Foundation of China under Grant 62522608, the Guangdong Provincial Project (Grant No. 2024TQ08X330), the Medical-Engineering Interdisciplinary Research Foundation of Shenzhen University under Grant 2024YG008, Brain Science and Brain-like Intelligence Technology - National Science and Technology Major Project (2021ZD0200500), Shenzhen-Hong Kong Institute of Brain Science-Shenzhen Fundamental Research Institutions (2023SHIBS0003), the Shenzhen Science and Technology Program (No. JCYJ20241202124222027 and JCYJ20241202124209011), and the Key Research and Development Program of Hunan Province (2025QK3008).


\ifCLASSOPTIONcaptionsoff
  \newpage
\fi

\bibliographystyle{IEEEtran}
\bibliography{references}

@article{he2020euclidean,
  author = {He, He and Wu, Dongrui},
  title = {Transfer Learning for Brain--Computer Interfaces: A {Euclidean} Space Data Alignment Approach},
  journal = {IEEE Trans. Biomed. Eng.},
  volume = {67},
  number = {2},
  pages = {399--410},
  year = {2020},
  doi = {10.1109/TBME.2019.2913914},
  url = {https://doi.org/10.1109/TBME.2019.2913914}
}

@inproceedings{sun2016coral,
  author = {Sun, Baochen and Feng, Jiashi and Saenko, Kate},
  title = {Return of Frustratingly Easy Domain Adaptation},
  booktitle = {Proc. AAAI Conf. Artif. Intell.},
  volume = {30},
  number = {1},
  pages = {2058--2065},
  year = {2016},
  doi = {10.1609/aaai.v30i1.10306},
  url = {https://doi.org/10.1609/aaai.v30i1.10306}
}

@article{oquab2024dinov2,
  author = {Oquab, Maxime and Darcet, Timoth{\'e}e and Moutakanni, Th{\'e}o and Vo, Huy and Szafraniec, Marc and Khalidov, Vasil and Fernandez, Pierre and Haziza, Daniel and Massa, Francisco and El-Nouby, Alaaeldin and Assran, Mahmoud and Ballas, Nicolas and Galuba, Wojciech and Howes, Russell and Huang, Po-Yao and Li, Shang-Wen and Misra, Ishan and Rabbat, Michael and Sharma, Vasu and Synnaeve, Gabriel and Xu, Hu and J{\'e}gou, Herv{\'e} and Mairal, Julien and Labatut, Patrick and Joulin, Armand and Bojanowski, Piotr},
  title = {{DINOv2}: Learning Robust Visual Features without Supervision},
  journal = {Trans. Mach. Learn. Res.},
  year = {2024},
  issn = {2835-8856},
  url = {https://openreview.net/forum?id=a68SUt6zFt}
}

@book{cochran1977sampling,
  author = {Cochran, William G.},
  title = {Sampling Techniques},
  edition = {3},
  publisher = {John Wiley \& Sons},
  address = {New York, NY, USA},
  year = {1977},
  isbn = {047116240X}
}

@inproceedings{james1961quadratic,
  author = {James, W. and Stein, Charles},
  title = {Estimation with Quadratic Loss},
  booktitle = {Proc. 4th Berkeley Symp. Math. Statist. Probab.},
  editor = {Neyman, Jerzy},
  volume = {1},
  pages = {361--379},
  publisher = {University of California Press},
  address = {Berkeley, CA, USA},
  year = {1961}
}

@article{schonemann1966procrustes,
  author = {Sch{\"o}nemann, Peter H.},
  title = {A Generalized Solution of the Orthogonal {Procrustes} Problem},
  journal = {Psychometrika},
  volume = {31},
  number = {1},
  pages = {1--10},
  year = {1966},
  doi = {10.1007/BF02289451},
  url = {https://doi.org/10.1007/BF02289451}
}

@article{holm1979sequential,
  author = {Holm, Sture},
  title = {A Simple Sequentially Rejective Multiple Test Procedure},
  journal = {Scand. J. Stat.},
  volume = {6},
  number = {2},
  pages = {65--70},
  year = {1979},
  url = {https://www.jstor.org/stable/4615733}
}

@article{efron1979bootstrap,
  author = {Efron, Bradley},
  title = {Bootstrap Methods: Another Look at the Jackknife},
  journal = {Ann. Statist.},
  volume = {7},
  number = {1},
  pages = {1--26},
  year = {1979},
  doi = {10.1214/aos/1176344552},
  url = {https://doi.org/10.1214/aos/1176344552}
}

@incollection{glynn2002controlvariates,
  author = {Glynn, Peter W. and Szechtman, Roberto},
  title = {Some New Perspectives on the Method of Control Variates},
  booktitle = {Monte Carlo and Quasi-Monte Carlo Methods 2000},
  pages = {27--49},
  publisher = {Springer},
  address = {Berlin, Heidelberg},
  year = {2002},
  doi = {10.1007/978-3-642-56046-0_3},
  url = {https://doi.org/10.1007/978-3-642-56046-0_3}
}

@inproceedings{CLIP,
  author = {Radford, Alec and Kim, Jong Wook and Hallacy, Chris and Ramesh, Aditya and Goh, Gabriel and Agarwal, Sandhini and Sastry, Girish and Askell, Amanda and Mishkin, Pamela and Clark, Jack and Krueger, Gretchen and Sutskever, Ilya},
  title = {Learning transferable visual models from natural language supervision},
  booktitle = {Proc. 38th Int. Conf. Mach. Learn.},
  series = {Proceedings of Machine Learning Research},
  volume = {139},
  pages = {8748--8763},
  publisher = {PMLR},
  year = {2021},
  url = {https://proceedings.mlr.press/v139/radford21a.html}
}

@article{Things-EEG2,
  author = {Gifford, Alessandro T and Dwivedi, Kshitij and Roig, Gemma and Cichy, Radoslaw M},
  title = {A large and rich {EEG} dataset for modeling human visual object recognition},
  journal = {NeuroImage},
  volume = {264},
  eid = {119754},
  publisher = {Elsevier},
  year = {2022},
  doi = {10.1016/j.neuroimage.2022.119754},
  note = {Art. no. 119754},
  url = {https://doi.org/10.1016/j.neuroimage.2022.119754}
}

@article{Things-data,
  author = {Hebart, Martin N and Contier, Oliver and Teichmann, Lina and Rockter, Adam H and Zheng, Charles Y and Kidder, Alexis and Corriveau, Anna and Vaziri-Pashkam, Maryam and Baker, Chris I},
  title = {{THINGS-data}, a multimodal collection of large-scale datasets for investigating object representations in human brain and behavior},
  journal = {eLife},
  volume = {12},
  eid = {e82580},
  publisher = {eLife Sciences Publications, Ltd},
  year = {2023},
  doi = {10.7554/eLife.82580},
  note = {Art. no. e82580},
  url = {https://doi.org/10.7554/eLife.82580}
}

@misc{xu2025alljoined16mmilliontrialeegimagedataset,
  author = {Jonathan Xu and Ugo Bruzadin Nunes and Wangshu Jiang and Samuel Ryther and Jordan Pringle and Paul S. Scotti and Arnaud Delorme and Reese Kneeland},
  title = {{Alljoined-1.6M}: A Million-Trial {EEG}-Image Dataset for Evaluating Affordable Brain-Computer Interfaces},
  year = {2025},
  eprint = {2508.18571},
  archivePrefix = {arXiv},
  primaryClass = {q-bio.NC},
  note = {arXiv:2508.18571},
  url = {https://arxiv.org/abs/2508.18571}
}

@misc{chenMindsEyeImage2024,
  author = {Chen, Chi-Sheng and Wei, Chun-Shu},
  title = {{Mind's Eye}: Image Recognition by {EEG} via Multimodal Similarity-Keeping Contrastive Learning},
  year = {2024},
  doi = {10.48550/arXiv.2406.16910},
  eprint = {2406.16910},
  archivePrefix = {arXiv},
  primaryClass = {eess.SP},
  note = {arXiv:2406.16910},
  url = {https://arxiv.org/abs/2406.16910}
}

@misc{jiangSubjectAwareMultiGranularityAlignment2026,
  author = {Jiang, Lin and She, Qingshan and Xu, Jiale and Xu, Haiqi and Wu, Duanpo and Kuang, Zhenzhong},
  title = {Subject-Aware Multi-Granularity Alignment for Zero-Shot {EEG}-to-Image Retrieval},
  year = {2026},
  doi = {10.48550/arXiv.2604.17782},
  eprint = {2604.17782},
  archivePrefix = {arXiv},
  primaryClass = {cs.CV},
  note = {arXiv:2604.17782},
  url = {https://arxiv.org/abs/2604.17782}
}

@article{joHyFIHyperbolicFeature2026,
  author = {Jo, Sangmin and Jeong, Wootaek and Heo, Da-Woon and Hwang, Yoohwan and Suk, Heung-Il},
  title = {{HyFI}: Hyperbolic Feature Interpolation for Brain-Vision Alignment},
  journal = {Proc. AAAI Conf. Artif. Intell.},
  volume = {40},
  number = {7},
  pages = {5575--5583},
  year = {2026},
  doi = {10.1609/aaai.v40i7.37476},
  url = {https://ojs.aaai.org/index.php/AAAI/article/view/37476}
}

@inproceedings{liNeuralMCRLNeuralMultimodal2024,
  author = {Li, Yueyang and Kang, Zijian and Gong, Shengyu and Dong, Wenhao and Zeng, Weiming and Yan, Hongjie and Siok, Wai Ting and Wang, Nizhuan},
  title = {Neural-{MCRL}: Neural Multimodal Contrastive Representation Learning for {EEG}-based Visual Decoding},
  booktitle = {Proc. IEEE Int. Conf. Multimedia Expo (ICME)},
  pages = {1--6},
  publisher = {IEEE},
  year = {2025},
  doi = {10.1109/ICME59968.2025.11210130},
  url = {https://doi.org/10.1109/ICME59968.2025.11210130}
}

@misc{linSIMONSaliencyawareIntegrative2026,
  author = {Lin, YuSheng and Tsai, Ji-Hwa and Wei, Chun-Shu},
  title = {{SIMON}: Saliency-aware Integrative Multi-view Object-centric Neural Decoding},
  year = {2026},
  doi = {10.48550/arXiv.2605.00401},
  eprint = {2605.00401},
  archivePrefix = {arXiv},
  primaryClass = {cs.CV},
  note = {arXiv:2605.00401},
  url = {https://arxiv.org/abs/2605.00401}
}

@inproceedings{liuLinguisticPriorsVisual2026,
  author = {Liu, Dongjun and Dai, Weichen and Qian, Jingsheng and Liu, Honggang and Yi, Hangjie and Kong, Wanzeng},
  title = {Linguistic Priors for Visual Decoupling: Towards Symmetric Vision-Brain Alignment},
  booktitle = {Proc. IEEE/CVF Conf. Comput. Vis. Pattern Recognit. (CVPR)},
  pages = {7869--7878},
  year = {2026},
  month = jun
}

@inproceedings{songNICE2024,
  author = {Song, Yonghao and Liu, Bingchuan and Li, Xiang and Shi, Nanlin and Wang, Yijun and Gao, Xiaorong},
  title = {Decoding natural images from {EEG} for object recognition},
  booktitle = {Proc. 12th Int. Conf. Learn. Represent.},
  year = {2024},
  url = {https://openreview.net/forum?id=dhLIno8FmH}
}

@inproceedings{liVisualDecodingReconstruction2024a,
  author = {Li, Dongyang and Wei, Chen and Li, Shiying and Zou, Jiachen and Liu, Quanying},
  title = {Visual Decoding and Reconstruction via {EEG} Embeddings with Guided Diffusion},
  booktitle = {Adv. Neural Inf. Process. Syst.},
  volume = {37},
  pages = {102822--102864},
  publisher = {Curran Associates, Inc.},
  year = {2024},
  doi = {10.52202/079017-3266},
  url = {https://doi.org/10.52202/079017-3266}
}

@misc{tangAligningWhatEEG2026,
  author = {Tang, Jingyi and Jiang, Shuai and Su, Fei and Zhao, Zhicheng},
  title = {Aligning What {EEG} Can See: Structural Representations for Brain-Vision Matching},
  year = {2026},
  doi = {10.48550/arXiv.2603.07077},
  eprint = {2603.07077},
  archivePrefix = {arXiv},
  primaryClass = {cs.CV},
  note = {arXiv:2603.07077},
  url = {https://arxiv.org/abs/2603.07077}
}

@misc{wangCORTIVACandidateScoreFusion2026,
  author = {Wang, Junhan and Chen, Kani},
  title = {{CORTIVA}: Candidate-Score Fusion of Complementary Visual Teachers for {EEG}- and {MEG}-to-Image Retrieval},
  year = {2026},
  doi = {10.48550/arXiv.2608.01355},
  eprint = {2608.01355},
  archivePrefix = {arXiv},
  primaryClass = {cs.CV},
  note = {arXiv:2608.01355},
  url = {https://arxiv.org/abs/2608.01355}
}

@misc{wangNeuroCLIPBrainInspiredPrompt2025a,
  author = {Wang, Jiyuan and Yao, Xizhang and Lin, Haipeng and Li, Ziyu and Liu, Yan and Liu, Qile and Huang, Gan and Zhang, Li and Liang, Zhen and Wu, Xia},
  title = {{NeuroCLIP}: Brain-Inspired Prompt Tuning for {EEG}-to-Image Multimodal Contrastive Learning},
  year = {2025},
  doi = {10.48550/arXiv.2511.09250},
  eprint = {2511.09250v2},
  archivePrefix = {arXiv},
  primaryClass = {cs.IR},
  note = {arXiv:2511.09250v2; Revised September 2026},
  url = {https://arxiv.org/abs/2511.09250v2}
}

@article{xuNeurovistaBidirectionalMasked2026,
  author = {Xu, Haopei and Lin, Ni},
  title = {{Neurovista}: A Bidirectional Masked Cross-Modal Fusion Network for Robust {EEG}-to-Image Decoding},
  journal = {Neural Netw.},
  volume = {195},
  eid = {108297},
  year = {2026},
  doi = {10.1016/j.neunet.2025.108297},
  note = {Art. no. 108297},
  url = {https://doi.org/10.1016/j.neunet.2025.108297}
}

@article{zhangNeuroBridgeBioInspiredSelfSupervised2025,
  author = {Zhang, Wenjiang and Wang, Sifeng and Su, Yuwei and Li, Xinyu and Zhang, Chen and Zhong, Suyu},
  title = {{NeuroBridge}: Bio-Inspired Self-Supervised {EEG}-to-Image Decoding via Cognitive Priors and Bidirectional Semantic Alignment},
  journal = {Proc. AAAI Conf. Artif. Intell.},
  volume = {40},
  number = {21},
  pages = {18028--18036},
  year = {2026},
  doi = {10.1609/aaai.v40i21.38863},
  url = {https://ojs.aaai.org/index.php/AAAI/article/view/38863}
}

@inproceedings{zhouEEGiTTeachingVision,
  author = {Zhou, Jiahao and Xu, Chenghao and Wang, Wei and Yang, Erkun and Deng, Cheng},
  title = {{EEGiT}: Teaching Vision Transformers to Understand the {EEG} Signal},
  booktitle = {Proc. IEEE/CVF Conf. Comput. Vis. Pattern Recognit. (CVPR)},
  pages = {40438--40447},
  year = {2026},
  month = jun
}

@inproceedings{zhuRegionAwareHierarchicalSubFeature,
  author = {Zhu, Yanan and Xiang, Ziwei and Wu, Jiamin and Guo, Jinyang and Zhang, Hongyuan and Song, Chunfeng and Fang, Hongjian and Guo, Yufei and Liu, Xianglong},
  title = {Region-Aware Hierarchical Sub-Feature Alignment for Robust {EEG}-Based Visual Decoding},
  booktitle = {Findings IEEE/CVF Conf. Comput. Vis. Pattern Recognit. (CVPR)},
  pages = {6667--6676},
  year = {2026},
  month = jun
}

@inproceedings{choiSubjectInvariantEEGEmbeddings2026,
  author = {Choi, Woohyeok and Kim, Jun-Mo and Kim, Songha and Choi, Yebin and Kam, Tae-Eui},
  title = {Subject-Invariant {EEG} Embeddings via Mixup and Adversarial Learning for Semantic Retrieval},
  booktitle = {Proc. 41st ACM/SIGAPP Symp. Appl. Comput.},
  pages = {1960--1961},
  publisher = {Association for Computing Machinery},
  year = {2026},
  doi = {10.1145/3748522.3779935},
  url = {https://doi.org/10.1145/3748522.3779935}
}

@inproceedings{huangNEEDCrossSubjectCrossTask2025,
  author = {Huang, Shuai and Luo, Huan and Jing, Haodong and Zhang, Qixian and Chang, Litao and Feng, Yating and Lin, Xiao and Qin, Chendong and Chen, Han and Jia, Shuwen and Sun, Siyi and Wang, Yongxiong},
  title = {{NEED}: Cross-Subject and Cross-Task Generalization for Video and Image Reconstruction from {EEG} Signals},
  booktitle = {Adv. Neural Inf. Process. Syst.},
  volume = {38},
  pages = {173134--173173},
  publisher = {Curran Associates, Inc.},
  year = {2025},
  doi = {10.52202/085713-5761},
  url = {https://doi.org/10.52202/085713-5761}
}

@inproceedings{kimCroSSCrosssubjectVisual2026b,
  author = {Kim, Jun-Mo and Choi, Woohyeok and Choi, Yebin and Kim, Songha and Kam, Tae-Eui},
  title = {{CroSS}: Cross-subject Visual Brain Decoding via Soft Alignment and Subject-aware Batch Normalization},
  booktitle = {Proc. 41st ACM/SIGAPP Symp. Appl. Comput.},
  pages = {1917--1924},
  publisher = {Association for Computing Machinery},
  year = {2026},
  doi = {10.1145/3748522.3779936},
  url = {https://doi.org/10.1145/3748522.3779936}
}

@article{liCrossSubjectGeneralizationEEG2026a,
  author = {Li, Taida and Yan, Yujun and Dou, Fei and Song, Wenzhan and Zhang, Xiang},
  title = {Cross-Subject Generalization for {EEG} Decoding: A Survey of Deep Learning Methods},
  journal = {Prog. Biomed. Eng.},
  volume = {8},
  number = {2},
  eid = {022013},
  year = {2026},
  doi = {10.1088/2516-1091/ae65f0},
  note = {Art. no. 022013},
  url = {https://doi.org/10.1088/2516-1091/ae65f0}
}

@article{luCausalityDrivenConvolutionalManifold2026a,
  author = {Lu, Bin and Chen, Junxiang and Wang, Fuwang and Wen, Guilin and Fu, Rongrong and Hua, Changchun},
  title = {Causality-Driven Convolutional Manifold Attention Network for Electroencephalogram Signal Decoding},
  journal = {IEEE Trans. Pattern Anal. Mach. Intell.},
  volume = {48},
  number = {3},
  pages = {2253--2264},
  year = {2026},
  doi = {10.1109/TPAMI.2025.3625631},
  url = {https://doi.org/10.1109/TPAMI.2025.3625631}
}

@misc{wangWhatCausesPerformance2026,
  author = {Wang, Yihe and Li, Taida and Yan, Yujun and Song, Wenzhan and Zhang, Xiang},
  title = {What Causes Performance Degradation in Cross-Subject {EEG} Classification?},
  year = {2024},
  doi = {10.48550/arXiv.2410.03057},
  eprint = {2410.03057},
  archivePrefix = {arXiv},
  primaryClass = {cs.CE},
  note = {arXiv:2410.03057; Accepted to IEEE International Conference on Systems, Man, and Cybernetics (SMC) 2026},
  url = {https://arxiv.org/abs/2410.03057}
}

@misc{zhuCrossPTEEGBenchmarkCrossParticipant2024,
  author = {Zhu, Shuqi and Ye, Ziyi and Ai, Qingyao and Liu, Yiqun},
  title = {{CrossPT-EEG}: A Benchmark for Cross-Participant and Cross-Time Generalization of {EEG}-based Visual Decoding},
  year = {2024},
  doi = {10.48550/arXiv.2406.07151},
  eprint = {2406.07151},
  archivePrefix = {arXiv},
  primaryClass = {cs.MM},
  note = {arXiv:2406.07151},
  url = {https://arxiv.org/abs/2406.07151}
}

@misc{cuiSCORESubjectCoordinate2026,
  author = {Cui, Zhenyao and Kan, Siyuan and Li, Siyang and Wang, Ziwei and Wu, Dongrui},
  title = {{SCORE}: Subject Coordinate Recovery for Label-Free Cross-Subject {EEG}-to-Image Retrieval},
  year = {2026},
  doi = {10.48550/arXiv.2608.19134},
  eprint = {2608.19134},
  archivePrefix = {arXiv},
  primaryClass = {cs.LG},
  note = {arXiv:2608.19134},
  url = {https://arxiv.org/abs/2608.19134}
}

@inproceedings{huangSATTCStructureAwareLabelFree,
  author = {Huang, Qunjie and Zhu, Weina},
  title = {{SATTC}: Structure-Aware Label-Free Test-Time Calibration for Cross-Subject {EEG}-to-Image Retrieval},
  booktitle = {Proc. IEEE/CVF Conf. Comput. Vis. Pattern Recognit. (CVPR)},
  pages = {16887--16896},
  year = {2026},
  month = jun
}

@article{jiangMindShotFewshotBrain2026,
  author = {Jiang, Shuai and Meng, Zhu and Li, Haiwen and Liu, Delong and Su, Fei and Zhao, Zhicheng},
  title = {{MindShot}: A Few-Shot Brain Decoding Framework via Transferring Cross-Subject Prior and Distilling Frequency Domain Knowledge},
  journal = {Knowl.-Based Syst.},
  volume = {332},
  eid = {114867},
  year = {2026},
  doi = {10.1016/j.knosys.2025.114867},
  note = {Art. no. 114867},
  url = {https://doi.org/10.1016/j.knosys.2025.114867}
}

@misc{kneelandENIGMAEEGtoImage152026,
  author = {Kneeland, Reese and Jiang, Wangshu and Nunes, Ugo Bruzadin and Scotti, Paul Steven and Delorme, Arnaud and Xu, Jonathan},
  title = {{ENIGMA}: {EEG}-to-Image in 15 Minutes Using Less Than 1\% of the Parameters},
  year = {2026},
  doi = {10.48550/arXiv.2602.10361},
  eprint = {2602.10361},
  archivePrefix = {arXiv},
  primaryClass = {q-bio.NC},
  note = {arXiv:2602.10361},
  url = {https://arxiv.org/abs/2602.10361}
}

@inproceedings{liuMindAdapterFewShotParameterEfficient2026,
  author = {Liu, Jiaxiang and Du, Jiawei and Chen, Xupeng and Li, Guoqi and Cai, Jiang and Fong, Simon and Xu, Mingkun},
  title = {{MindAdapter}: Few-Shot Parameter-Efficient Residual Calibration of Cross-Subject Brain-to-Visual Decoding Models},
  booktitle = {Proc. 32nd ACM SIGKDD Conf. Knowl. Discov. Data Min. V.2},
  pages = {11456--11467},
  publisher = {Association for Computing Machinery},
  year = {2026},
  doi = {10.1145/3770855.3818949},
  url = {https://dl.acm.org/doi/10.1145/3770855.3818949}
}

@inproceedings{daiMindAlignerExplicitBrain2025,
  author = {Dai, Yuqin and Yao, Zhouheng and Song, Chunfeng and Zheng, Qihao and Mai, Weijian and Peng, Kunyu and Lu, Shuai and Ouyang, Wanli and Yang, Jian and Wu, Jiamin},
  title = {{MindAligner}: Explicit Brain Functional Alignment for Cross-Subject Visual Decoding from Limited {fMRI} Data},
  booktitle = {Proc. 42nd Int. Conf. Mach. Learn.},
  series = {Proceedings of Machine Learning Research},
  volume = {267},
  pages = {12214--12228},
  publisher = {PMLR},
  year = {2025},
  url = {https://proceedings.mlr.press/v267/dai25m.html}
}

@article{ferranteTheirEyesMultisubject2024a,
  author = {Ferrante, Matteo and Boccato, Tommaso and Ozcelik, Furkan and VanRullen, Rufin and Toschi, Nicola},
  title = {Through Their Eyes: Multi-subject Brain Decoding with Simple Alignment Techniques},
  journal = {Imaging Neurosci.},
  volume = {2},
  pages = {1--21},
  year = {2024},
  doi = {10.1162/imag_a_00170},
  url = {https://doi.org/10.1162/imag_a_00170}
}

@article{haxbyCommonHighDimensionalModel2011,
  author = {Haxby, James V. and Guntupalli, J. Swaroop and Connolly, Andrew C. and Halchenko, Yaroslav O. and Conroy, Bryan R. and Gobbini, M. Ida and Hanke, Michael and Ramadge, Peter J.},
  title = {A Common, High-Dimensional Model of the Representational Space in Human Ventral Temporal Cortex},
  journal = {Neuron},
  volume = {72},
  number = {2},
  pages = {404--416},
  year = {2011},
  doi = {10.1016/j.neuron.2011.08.026},
  url = {https://doi.org/10.1016/j.neuron.2011.08.026}
}

@article{rodriguesRiemannianProcrustesAnalysis2019a,
  author = {Rodrigues, Pedro Luiz Coelho and Jutten, Christian and Congedo, Marco},
  title = {{Riemannian} {Procrustes} Analysis: Transfer Learning for Brain--Computer Interfaces},
  journal = {IEEE Trans. Biomed. Eng.},
  volume = {66},
  number = {8},
  pages = {2390--2401},
  year = {2019},
  doi = {10.1109/TBME.2018.2889705},
  url = {https://doi.org/10.1109/TBME.2018.2889705}
}

@inproceedings{scottiMindEye2SharedSubjectModels2024,
  author = {Scotti, Paul Steven and Tripathy, Mihir and Torrico, Cesar and Kneeland, Reese and Chen, Tong and Narang, Ashutosh and Santhirasegaran, Charan and Xu, Jonathan and Naselaris, Thomas and Norman, Kenneth A. and Abraham, Tanishq Mathew},
  title = {{MindEye2}: Shared-Subject Models Enable {fMRI}-To-Image With 1 Hour of Data},
  booktitle = {Proc. 41st Int. Conf. Mach. Learn.},
  series = {Proceedings of Machine Learning Research},
  volume = {235},
  pages = {44038--44059},
  publisher = {PMLR},
  year = {2024},
  url = {https://proceedings.mlr.press/v235/scotti24a.html}
}

@article{horikawaGenericDecodingSeen2017,
  author = {Horikawa, Tomoyasu and Kamitani, Yukiyasu},
  title = {Generic Decoding of Seen and Imagined Objects Using Hierarchical Visual Features},
  journal = {Nat. Commun.},
  volume = {8},
  number = {1},
  eid = {15037},
  year = {2017},
  doi = {10.1038/ncomms15037},
  note = {Art. no. 15037},
  url = {https://doi.org/10.1038/ncomms15037}
}

@article{palazzoDecodingBrainRepresentations2021,
  author = {Palazzo, Simone and Spampinato, Concetto and Kavasidis, Isaak and Giordano, Daniela and Schmidt, Joseph and Shah, Mubarak},
  title = {Decoding Brain Representations by Multimodal Learning of Neural Activity and Visual Features},
  journal = {IEEE Trans. Pattern Anal. Mach. Intell.},
  volume = {43},
  number = {11},
  pages = {3833--3849},
  year = {2021},
  doi = {10.1109/TPAMI.2020.2995909},
  url = {https://doi.org/10.1109/TPAMI.2020.2995909}
}

@article{kamitani2005decoding,
  author = {Kamitani, Yukiyasu and Tong, Frank},
  title = {Decoding the visual and subjective contents of the human brain},
  journal = {Nat. Neurosci.},
  volume = {8},
  number = {5},
  pages = {679--685},
  publisher = {Nature Publishing Group US New York},
  year = {2005},
  doi = {10.1038/nn1444},
  url = {https://doi.org/10.1038/nn1444}
}

@article{kay2008identifying,
  author = {Kay, Kendrick N and Naselaris, Thomas and Prenger, Ryan J and Gallant, Jack L},
  title = {Identifying natural images from human brain activity},
  journal = {Nature},
  volume = {452},
  number = {7185},
  pages = {352--355},
  publisher = {Nature Publishing Group UK London},
  year = {2008},
  doi = {10.1038/nature06713},
  url = {https://doi.org/10.1038/nature06713}
}

@article{naselaris2009bayesian,
  author = {Naselaris, Thomas and Prenger, Ryan J and Kay, Kendrick N and Oliver, Michael and Gallant, Jack L},
  title = {Bayesian reconstruction of natural images from human brain activity},
  journal = {Neuron},
  volume = {63},
  number = {6},
  pages = {902--915},
  publisher = {Elsevier},
  year = {2009},
  doi = {10.1016/j.neuron.2009.09.006},
  url = {https://doi.org/10.1016/j.neuron.2009.09.006}
}

\end{document}